\documentclass[10pt]{article}

\newif\ifpreprint
\preprinttrue
\ifdefined\ANONYMOUS \preprintfalse \fi

\ifpreprint
  \usepackage[preprint]{tmlr}
\else
  \usepackage{tmlr}
\fi

\usepackage{amsmath,amssymb,amsthm}
\usepackage{booktabs}
\usepackage{array}
\usepackage{float}
\usepackage{multirow}
\usepackage{makecell}
\usepackage{graphicx}
\usepackage{epstopdf}  
\usepackage{xcolor}
\usepackage{algorithm}
\usepackage[noend]{algpseudocode}
\usepackage{microtype}
\usepackage[hidelinks]{hyperref}
\usepackage{url}
\usepackage[capitalise,nameinlink]{cleveref}
\crefname{figure}{Figure}{Figures}
\Crefname{figure}{Figure}{Figures}
\newcolumntype{L}[1]{>{\raggedright\arraybackslash}p{#1}}
\newcommand{\Eccd}{E_{\mathrm{CCD}}}
\newcommand{\Sph}{\mathbb{S}^2}
\newcommand{\Hess}{\operatorname{Hess}}

\ifpreprint
  \newcommand{\repo}{\url{https://github.com/jesse87wang-bit/thomson-records}}
  \newcommand{\zenodo}{\url{https://doi.org/10.5281/zenodo.23996279}}
\else
  \newcommand{\repo}{[anonymised repository]}
  \newcommand{\zenodo}{[anonymised archive]}
\fi

\title{Certified Improvements to Best-Known Thomson-Problem\\ Configurations with a Search Pipeline Built with\\ Large Language Model Agents}

\author{\name Hui Shao\thanks{Joint first authors.} \\
        \addr Zhejiang University
        \AND
        \name Tao Wang\footnotemark[1] \\
        \addr Independent Researcher
        \AND
        \name Xiaotian Liu \\
        \addr Georgia Institute of Technology
        \AND
        \name Yijie Peng\thanks{Corresponding author.} \\
        \addr Nanjing University}

\def\month{11}
\def\year{2026}
\def\openreview{\url{https://openreview.net/forum?id=XXXX}}

\begin{document}

\maketitle

\begin{abstract}
The Thomson problem asks for $N$ unit charges on a sphere with minimum Coulomb energy; for large $N$ the best configurations known are curated in the Cambridge Cluster Database, whose largest-$N$ entries were set in October 2026. We report configurations for $N=912$, $1382$ and $2502$ with energies $0.0036186$, $0.0109730$ and $0.0614033$ below the database values, certified by exact integer arithmetic. All three are numerically strict local minima modulo rotations and keep the defect inventory and scar compositions of the database configurations; the database value for $N=912$ had been reached by two different methods. Our search runs unconstrained replica-exchange basin hopping from the database configurations, with a shared pool of minima, recombination of improved regions after relabelling by geometry, and gradient-only polishing; replica-exchange basin hopping alone found the $N=912$ and $1382$ configurations, the former in one of five runs. We built and ran it with large language model agents (Claude, through a chat interface and the Claude Code agent), and we report every search, including those that failed. Two findings matter for others who search the database. Polishing alone lowers two entries by more than the threshold at which our pipeline reports a candidate, which produced two spurious records until candidates were compared with the polished database energies (the polished baseline). A local-relaxation speed-up passed its exactness test yet found nothing in a controlled comparison in which every arm with full relaxation went below the database value within 15 minutes. We distil three practices for computational discovery with large language model agents: calibrate the search code and verify candidates with code separate from the search, compare candidates with the polished baseline, and test design changes in controlled comparisons with pass criteria fixed in advance.
\end{abstract}

\section{Introduction}
\label{sec:intro}

Place $N$ unit charges on the unit sphere $\Sph\subset\mathbb{R}^3$ so that their Coulomb energy
\begin{equation}
  E(x_1,\cdots,x_N)\;=\;\sum_{1\le i<j\le N}\frac{1}{\lVert x_i-x_j\rVert}
  \label{eq:energy}
\end{equation}
is as small as possible. This is the Thomson problem \citep{thomson1904}. Its global minimum has been proved only for a handful of small $N$ \citep{schwartz2013,cohn2007}; the closely related task of computing near-optimal points for the logarithmic energy is Smale's seventh problem \citep{smale1998,saff1997,borodachov2019}, and the minimisers are models of spherical crystals whose topologically required defects organise into ``scars'' of dislocations \citep{perezgarrido1997,bowick2000,bowick2002,bausch2003,wales2006thomson,wales2009defect}. Because the number of local minima appears to grow exponentially with $N$ \citep{erber1991,amore2025landscape}, the problem is also a long-standing benchmark for global optimisation \citep{wales2003book}. For large $N$ the best configurations known are curated in the Cambridge Cluster Database \citep[CCD;][]{ccd}. Its 53 Coulomb entries span $512\le N\le 9913$; the current values for the largest sizes, including $N=1382$ and $N=2502$, were set by a search method published in October 2026 \citep[hereafter the October 2026 method;][]{amore2026efficient}, which substantially lowered earlier basin-hopping results \citep{wales2009defect}.

At the same time, artificial intelligence (AI) systems are increasingly used to search for mathematical constructions, from reinforcement learning and neural-guided search \citep{wagner2021,fawzi2022alphatensor,charton2024patternboost} to evolutionary search over programs written by large language models (LLMs) \citep{romera2024funsearch,novikov2025alphaevolve}. AlphaEvolve was applied to the Thomson problem itself: for $N$ up to about $300$, a range searched intensively for decades, it reproduced the best-known energies to high precision but did not improve them \citep{georgiev2025}. LLMs and agent teams have also contributed to machine-checked proofs of optimality for $N=7$ and $N=8$ \citep{tran2026lean,kryvonos2026eight,toobysmith2026warrant}. These systems are engineered pipelines in which the model proposes and an automated evaluator scores. A different and widely used mode is a general-purpose LLM assistant or coding agent \citep{jimenez2024swebench,anthropic2025claudecode} that writes and runs research code while a human decides what to try next. This mode has already touched the CCD: in October 2026, within a day of the problem being posted in a public AI research challenge, two participants working with the Claude Code agent posted configurations up to $0.045$ below the CCD value for $N=960$ \citep{fewshots2026,abreto2026p007,ronnie2025p007}. Such results are reported as leaderboard entries; how reliable this way of working is, and what it takes to turn its output into a claim others can trust, is rarely documented.

We used this mode of working to search the CCD for improvable entries. Claude, used through a chat interface (hereafter the chat assistant), designed and wrote the search and verification code, and the Claude Code agent running on a laptop (hereafter the agent) deployed, calibrated, ran and patched it; we chose the problem and the sizes, set time boxes and stop rules, judged the results and decided what to release (\cref{tab:roles}). Where it matters who acted, we name the chat assistant or the agent; ``the AI systems'' means both. Our contributions are as follows.
\begin{enumerate}
  \item We give configurations for $N=912$, $1382$ and $2502$ with energies $0.0036186$, $0.0109730$ and $0.0614033$ below the CCD values, found by a search pipeline that runs unconstrained replica-exchange basin hopping from the CCD configurations with a shared pool, recombination of improved regions and gradient-only polishing (\cref{sec:method,sec:results}); replica-exchange basin hopping alone, without pool or recombination, found the configurations for $N=912$ and $N=1382$. The improvements are certified by exact integer arithmetic (\cref{sec:cert}), exceed by a factor of more than $10^{6}$ the $10^{-9}$ agreement between our floating-point evaluations, and were reproduced by a floating-point recomputation written for this paper (hereafter the paper's verification code), which shares no code with the search. Numerically, all three configurations are strict local minima modulo rotations with the defect inventory and scar compositions of the CCD configurations, so for these sizes the CCD configurations were not the lowest minima with their defect inventory and scar compositions. The CCD value for $N=912$ predates the October 2026 method and was reproduced by it.
  \item We show that comparison with the CCD values is not enough: polishing alone lowers two of the 53 CCD entries by more than the $10^{-5}$ threshold at which our pipeline reports a candidate (hereafter the candidate threshold), which produced two spurious records before candidates were compared with the lower of the CCD value and the energy of the polished CCD configuration (hereafter the polished baseline; \cref{sec:dividend}).
  \item We show, in a controlled comparison whose pass criterion was fixed in advance (according to the project notes), that a local-relaxation speed-up designed by the chat assistant, exact to $10^{-9}$ in its energy accounting, found no configuration below the CCD value in four core-hours, while every arm that included full relaxation did so within $15$ wall-clock minutes; measurements on identical moves show that local relaxation leaves the starting minimum less often and, when it does leave, reports energies $0.1$ to $0.4$ above those of full relaxation from the same move (\cref{sec:speedup}).
\end{enumerate}
Every search we ran, including the thirteen sizes that did not improve, is reported with its budget (\cref{sec:method,sec:results}, \cref{app:sweep,app:probe}), together with the division of labour between us and the AI systems. From these results we draw three practices for AI-assisted computational discovery (\cref{sec:discussion}).

\paragraph{Scope of the claims.} We claim energies lower than the CCD values as listed on 6 and 10 November 2026 (for $N=912$, on 10 November); to our knowledge no lower energies for these sizes had been published by 10 November 2026 (the CCD page and files were unchanged, the arXiv version of \citet{amore2026efficient} had not been revised, and we found no other report). We do not claim global optimality. We do not claim that our search method is better than that of \citet{amore2026efficient}: our search on $N=2502$ used about $12$ core-hours before the final configuration appeared ($10$ to $100$ times the per-size timings they report) and about $58$ in total, and we ran no equal-budget comparison (\cref{sec:limitations}). In absolute terms the decreases ($0.0036$, $0.011$ and $0.061$ for $N=912$, $1382$ and $2502$) are comparable to or larger than the energy differences between the distinct low-lying minima our searches found near the CCD configurations ($10^{-5}$ to $10^{-2}$); in relative terms they are about $10^{-8}$. All three exceed by many orders of magnitude the $10^{-9}$ agreement between our floating-point evaluations and the width of the certified enclosures (below $3\times10^{-18}$; \cref{sec:cert}). We directed the project throughout; the search that produced the $N=912$ configuration was proposed and run by Claude Code in a later session, after the project, with our approval (\cref{app:probe}). In neither case did an AI system choose the problem or decide what to report.

\section{Related work}
\label{sec:related}

\paragraph{The Thomson problem and spherical crystals.} Numerical studies of the Thomson problem go back decades \citep{erber1991,altschuler1997}. For large $N$ the twelve disclinations required by Euler's formula are screened by chains of dislocations (scars) whose length grows with $N$ \citep{perezgarrido1997,perezgarrido1999,bowick2000,bowick2002}, as observed in colloidal crystals \citep{bausch2003}. Systematic surveys of low-energy configurations and their defect motifs were carried out by \citet{wales2006thomson} and \citet{wales2009defect}; icosahedral and slightly broken icosahedral constructions are discussed by \citet{altschuler2006} and \citet{roshal2015}. \citet{amore2026efficient} introduced oscillating-size hybrid basin hopping, which grows and shrinks configurations one charge at a time across a window of $N$, and used it to set the current CCD values for the largest sizes, including $N=1382$ and $N=2502$, reporting a few hundred to a few thousand core-seconds per size for the comparisons with earlier results; they note that improved configurations were still being found as their runs continued.

\paragraph{Global optimisation of clusters.} Basin hopping \citep{li1987,wales1997bh} performs a Monte Carlo walk on the set of local minima. Replica exchange (parallel tempering) runs several replicas at different temperatures and swaps them \citep{swendsen1986,hukushima1996,earl2005}; multi-funnel landscapes such as that of the 38-atom Lennard-Jones cluster \citep{doye1999lj38} motivate its use. Genetic algorithms with ``cut-and-splice'' crossover combine halves of parent clusters \citep{deaven1995} and were applied to the Thomson problem soon after \citep{morris1996thomsonga}. Our recombination step is closest to cut-and-splice but splices the region in which a parent differs from a common reference configuration, after the kind of permutational alignment that is routine for atomic clusters \citep{griffiths2017alignment,gmin}.

\paragraph{AI for mathematical and scientific discovery.} Neural or LLM-guided search has produced new constructions in combinatorics and algorithms \citep{wagner2021,davies2021,fawzi2022alphatensor,romera2024funsearch,charton2024patternboost,novikov2025alphaevolve,georgiev2025}; LLMs have been used directly as optimisers \citep{yang2024opro}; and agentic systems aim to automate larger parts of research \citep{lu2024aiscientist,gottweis2025coscientist}, with benchmarks that measure how well agents carry out machine-learning engineering, AI research and development, replication of published research, and data-driven science \citep{chan2024mlebench,wijk2024rebench,starace2025paperbench,chen2024scienceagentbench}. Work on the Thomson problem itself is summarised in \cref{sec:intro}. Open platforms now let AI agents compete on verified problems \citep{bianchi2026harnessing,fewshots2026}. The closest prior work is the pair of $N=960$ entries described in \cref{sec:intro} (\cref{app:n960}). Our work differs from these in that it targets mostly larger sizes, characterises the new configurations (local optimality and defect structure) beyond the exact energy check that the challenge's official verifier also performs, and documents the whole project, including the division of labour between us and the AI systems and how each failure was detected. As at $N=960$, the CCD values for $N=1382$ and $N=2502$ were recent; the $N=960$ entries achieved a larger relative decrease (about $10^{-7}$ against $1$--$2\times10^{-8}$). The emphasis on pass criteria fixed in advance follows practice in empirical science and machine learning \citep{nosek2018prereg,henderson2018}.

\section{Problem, certification, and local optimality}
\label{sec:cert}

A configuration $X=(x_1,\cdots,x_N)\in(\Sph)^N$ has energy \eqref{eq:energy}. The energy $E$ is invariant under rotations and reflections of the sphere and under relabelling of the points. We write $\Eccd(N)$ for the energy listed by the CCD (the CCD value) and $\ell=\sqrt{4\pi/N}$ for the length scale used throughout ($\ell=0.0709$ for $N=2502$; in the new and the CCD configurations for the three sizes we report the smallest pairwise distance is about $0.96\,\ell$). We report energies relative to the CCD value, $\Delta E=E-\Eccd(N)$, call $-\Delta E$ the decrease when it is positive and $-\Delta E/\Eccd(N)$ the relative decrease, and refer to configurations by $\Delta E$ (``the $-0.0569$ configuration''). We write $g=g(X)$ for the Riemannian (tangential) gradient of $E$ on $(\Sph)^N$, whose $i$th block is $(I_3-x_ix_i^{\top})\nabla_iE$, where $I_3$ is the $3\times3$ identity matrix and $\nabla_iE$ is the Euclidean gradient of $E$ with respect to $x_i$, and $|g|_\infty$ for the largest absolute value of its entries.

\paragraph{What counts as an improvement.} Floating-point evaluation of \eqref{eq:energy} for $N=2502$ sums $3.1\times10^6$ terms to a total of about $3\times10^6$, adjacent double-precision numbers near this value are $4.7\times10^{-10}$ apart (one unit in the last place, ulp), and a plain sum of so many terms can accumulate much larger rounding errors. The decreases we report exceed both by orders of magnitude, but we make the claim independent of floating point altogether.

Let $\tilde x_i=(a_i,b_i,c_i)/d_i$ with integers satisfying $a_i^2+b_i^2+c_i^2=d_i^2$, so that $\tilde x_i\in\Sph$ exactly, write $\tilde X=(\tilde x_1,\cdots,\tilde x_N)$, and let $K_{ij}=d_id_j-(a_ia_j+b_ib_j+c_ic_j)>0$. Since $\lVert\tilde x_i-\tilde x_j\rVert^2=2K_{ij}/(d_id_j)$, every term of $E(\tilde X)$ equals $\sqrt{d_id_j/(2K_{ij})}$, so for any integer $s>0$ the bounds
\begin{equation}
  L\;=\;\frac{1}{s}\sum_{i<j}\Big\lfloor \sqrt{\tfrac{s^2d_id_j}{2K_{ij}}}\Big\rfloor
  \;\le\; E(\tilde X)\;\le\;
  U\;=\;\frac{1}{s}\sum_{i<j}\Big\lceil \sqrt{\tfrac{s^2d_id_j}{2K_{ij}}}\Big\rceil
  \label{eq:cert}
\end{equation}
hold term by term. Both sums are evaluated exactly in integer arithmetic, because the floor and the ceiling of the square root of a rational number can be computed exactly with integers; we call \eqref{eq:cert} the integer certificate.

We map each point, given in floating point, to a rational point exactly on the sphere by rounding its stereographic coordinates (projecting from the pole opposite its largest coordinate) to the grid $2^{-29}\mathbb{Z}^2$ and inverting the projection; this moves points by at most $2.6\times10^{-9}$. With $s=2^{80}$ each term is bounded to within $2^{-80}$, so the enclosure \eqref{eq:cert} is narrower than $3\times10^{-18}$ for $N=2502$. We call an improvement certified if the upper bound $U$ for the rationalised new configuration is below $\Eccd(N)$; the rationalised configuration is itself a valid configuration, so no rounding argument is needed.

\paragraph{Local optimality.} The claim above does not require local optimality, but we check it to characterise the configurations. With $\nabla^2_{ij}E$ the $3\times3$ block of Euclidean second derivatives of $E$ with respect to $x_i$ and $x_j$, the Riemannian Hessian on $(\Sph)^N$ acts on a tangent vector $v=(v_1,\cdots,v_N)$ by \citep{absil2008}
\begin{equation}
  \big(\Hess E[v]\big)_i \;=\;(I_3-x_ix_i^{\top})\sum_{j}\nabla^2_{ij}E\,v_j\;-\;\big(x_i^{\top}\nabla_iE\big)\,v_i .
  \label{eq:hess}
\end{equation}
Because $E$ is rotation invariant, $\Hess E$ has a kernel of dimension at least three at any critical point. We call $X$ a strict local minimum modulo rotations if it is a critical point and the remaining $2N-3$ eigenvalues are positive. Our configurations are critical points to within $|g|_\infty<10^{-8}$, so this is a numerical statement; it could be made rigorous by an a-posteriori Newton--Kantorovich argument in interval arithmetic, which our main claim does not need. We assemble the $2N\times 2N$ matrix of \eqref{eq:hess} in orthonormal tangent bases, check it against finite differences of the Riemannian gradient (relative error about $2\times10^{-9}$), and compute its full spectrum.

\paragraph{Terminology.} A relaxation is a local minimisation of $E$ from a perturbed configuration; a full relaxation minimises over all $N$ points, and a local relaxation (\cref{sec:speedup}) only over the points nearest the perturbation. Polishing is a much tighter minimisation, to $|g|_\infty<10^{-8}$ (\cref{sec:method}). In a low-energy configuration most points have six neighbours in the Delaunay triangulation (coordination number six); points with five or seven are defects, and Euler's formula forces the number of 5-fold points to exceed that of 7-fold points by twelve. A bound 5--7 pair is a dislocation; a chain of alternating 5- and 7-fold points with one extra 5-fold point is a scar; the defect inventory of a configuration is its numbers of 5- and 7-fold points. Point groups are in Schoenflies notation ($C_1$: no symmetry; $C_n$: one $n$-fold axis; $D_n$: an $n$-fold axis and $n$ two-fold axes perpendicular to it; $T$: the rotations of a regular tetrahedron).

\section{Search pipeline and workflow}
\label{sec:method}

\begin{figure}[t]
\centering
\includegraphics{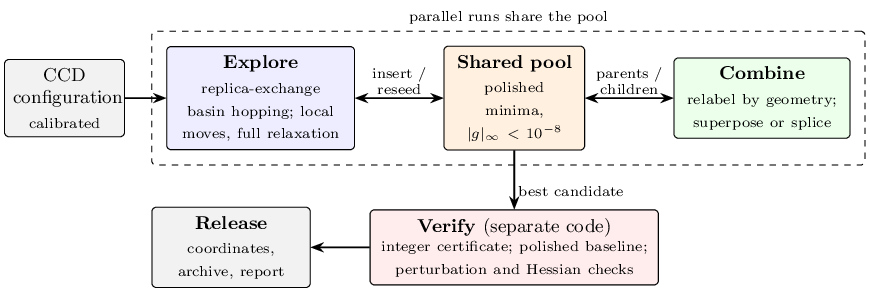}
\caption{The pipeline that produced the $N=2502$ configuration, with verification as finally applied (the comparison with the polished baseline was added after the last search stage, \cref{sec:dividend}; the Hessian analysis was done for this paper). We chose which sizes to run, for how long, and when to stop; the code was written by the chat assistant and deployed, run and patched by the agent (\cref{tab:roles}).}
\label{fig:pipeline}
\end{figure}

\Cref{fig:pipeline} summarises the pipeline; full parameters are in \cref{app:params}. The search ran in three stages, detailed at the end of this section: a two-hour screening of four sizes (stage~1), a $9.6$-hour search with a shared pool and recombination, mainly on the two sizes that improved (stage~2), and time slices over 11 sizes (stage~3). Each run at a new size starts from the CCD configuration; later stages restart from the best configurations in the shared pool.

\paragraph{Exploration: unconstrained replica-exchange basin hopping.} The search is unconstrained: no symmetry is imposed and all $3N$ coordinates move. A run is one single-core search process. Each run keeps three replicas at temperatures $0.003$, $0.0173$ and $0.1$ (energy units; geometrically spaced). A step perturbs each replica with a localised move centred on a random point (the move centre): (i) Gaussian displacements of standard deviation $\sigma\sim\mathcal{U}(0.1,0.35)\,\ell$, where $\mathcal{U}$ denotes a uniform distribution, applied to the $k\sim\mathcal{U}\{6,\cdots,29\}$ nearest neighbours of the move centre; (ii) a rotation by $\theta\sim\mathcal{U}(-0.35,0.35)$ rad of the $k\sim\mathcal{U}\{10,\cdots,39\}$ nearest neighbours about the axis through the move centre, which tends to make dislocations glide; or (iii) from stage~2 on, Gaussian displacements with $\sigma=0.08\,\ell$ of $k\sim\mathcal{U}\{40,\cdots,89\}$ neighbours. The perturbed configuration is then relaxed in all $3N$ coordinates by the limited-memory Broyden--Fletcher--Goldfarb--Shanno quasi-Newton method (L-BFGS) \citep{liu1989,byrd1995} on unnormalised coordinates $z_i$ with $x_i=z_i/\lVert z_i\rVert$, and accepted with the Metropolis rule at the replica's temperature. After each step, neighbouring replicas are swapped with the usual replica-exchange probability \citep{hukushima1996}. A relaxed configuration whose energy falls below the lowest energy in the shared pool (the pool best) by more than the polishing trigger of \cref{tab:params} is polished and, if new, inserted into the pool. Energies stored in the pool and compared with the baseline (the CCD energy recomputed from the posted coordinates, later corrected as described in \cref{sec:dividend}) are computed with compensated summation \citep{kahan1965,neumaier1974}; the fast plain sums are used only inside the walk.

\paragraph{Polishing.} At $E\approx3\times10^6$ the energy changes near a minimum fall below the resolution of an energy-based line search, and L-BFGS stalls with $|g|_\infty$ around $10^{-4}$. We therefore polish with a gradient-only Barzilai--Borwein iteration \citep{barzilai1988} on $(\Sph)^N$, $X_{t+1}=\mathrm{normalise}(X_t-\alpha_t\,g(X_t))$ with $\alpha_t=\lVert\Delta X_t\rVert^2/(\Delta X_t^{\top}\Delta g_t)$, $\Delta X_t=X_t-X_{t-1}$ and $\Delta g_t=g(X_t)-g(X_{t-1})$, where $\mathrm{normalise}$ maps each point $x_i$ to $x_i/\lVert x_i\rVert$, until $|g|_\infty<10^{-8}$ (stage-specific variants are listed in \cref{tab:params}).

\paragraph{Shared pool.} Runs exchange configurations through a shared pool of polished minima; a configuration enters only if its energy differs from that of every pool member by more than $2\times10^{-7}$. Every 30 minutes each run restarts its coldest replica from the pool best. A pool configuration more than the candidate threshold ($10^{-5}$) below the baseline is reported as a candidate.

\paragraph{Recombination.} Independent runs often improve different parts of the sphere. Combining two such configurations requires a point-to-point correspondence, but during a long walk identical points exchange places (a local rotation move, for example, permutes the points of a patch), so labels stop corresponding even where the geometry has barely changed: the stage-1 $-0.0149$ configuration, whose changes are confined to three small clusters of $22$ points, has an aligned fraction (the share of points within $0.5\,\ell$ of the same-numbered point of the CCD configuration) of only $0.51$. A separate process, the combiner, applies \cref{alg:combine} every five minutes to pairs of the ten best pool configurations below the baseline; \cref{alg:combine} relabels each configuration by geometry before combining (an early version compared raw labels; \cref{sec:failures}). Each child, which we call a combination, is relaxed, polished, and inserted if it is lower than both parents.

\begin{algorithm}[t]
\caption{Recombination of two improved configurations $A,B$ relative to the reference configuration $R$ (the CCD configuration)}
\label{alg:combine}
\begin{algorithmic}[1]
\State Align $A$ and $B$ to $R$: iterate an orthogonal Procrustes fit \citep{kabsch1976} with nearest-neighbour correspondences, then relabel points by nearest neighbour, resolving conflicts by a minimum-cost assignment \citep{kuhn1955}
\State Keep a configuration only if its aligned fraction relative to $R$ exceeds $0.9$
\State $m_A\gets\{i:\lVert A_i-R_i\rVert>0.15\,\ell\}$, enlarged by every point within $1.6\,\ell$ of a point in this set;\quad $m_B$ likewise
\State Form two children: $A$ with the points in $m_B$ replaced by those of $B$, and $B$ with the points in $m_A$ replaced by those of $A$ \Comment{region splices}
\If{$|m_A\cap m_B|<\tfrac12\min(|m_A|,|m_B|)$} \Comment{changes roughly disjoint}
  \State Form a third child, $\mathrm{normalise}\big(R+(A-R)+(B-R)\big)$ \Comment{displacement superposition}
\EndIf
\State Relax and polish each child; insert it into the pool if it is lower than both parents
\end{algorithmic}
\end{algorithm}

\paragraph{Stages and choice of sizes.} The project began with a ten-hour attempt at $N=960$ (the $N=960$ attempt; nine runs, about $80$ core-hours), starting from the best configuration then posted in the public challenge mentioned above; it found no improvement (\cref{app:n960}). This, together with the observation that the CCD values for the largest sizes were recent and that \citet{amore2026efficient} reported modest per-size timings, motivated a change of target: rather than ask how to search faster, we asked where further search was most likely to pay off. Stage~1 was a two-hour screening of four sizes ($N=1382$, $2000$, $2502$, $3552$; two unconstrained runs each, plus one symmetry-constrained run at $N=2502$ to try out the fallback) with a decision rule fixed in advance (according to the project notes): concentrate compute on any size that produced a configuration below the CCD value, and switch to symmetry-constrained searches if none did. Stage~2 ran the pool and the combiner for $9.6$ hours, in two segments, on $N=1382$ and $N=2502$ and searched three further sizes with one run each. Stage~3 cycled through 11 sizes whose CCD configurations have no symmetry ($C_1$) in time slices, and a controlled comparison on $N=2502$ then tested the design change made for stage~3 (\cref{sec:speedup}). The machine was a single ten-core laptop (\cref{tab:params}) running up to nine single-core processes (runs and the combiner). After the project we ran the stage-1 protocol, five runs per size, on the four CCD entries without symmetry just below $N=960$ ($N=900$, $912$, $920$ and $932$; the probe of smaller sizes, \cref{app:probe}).

\paragraph{Human--AI workflow.} \Cref{tab:roles} lists the division of labour as recorded in the project notes (the plans, status reports, decisions and summaries written during the project). Each stage started with the chat assistant, which designed the plan, wrote the code and delivered it as a package together with a written instruction for the agent. The agent installed the package, calibrated it against known energies, launched and monitored the runs, patched bugs (including, with our approval, the combiner's relabelling step), ran the verification scripts, wrote the status reports, and published the results. We read the reports, discussed them with the chat assistant and decided the next step. The project took about three and a half days (5--8 November 2026). A further AI system, used separately from the Claude sessions and not identified in the project notes (hereafter the reviewing AI system), reviewed the plan before the $N=960$ attempt and raised nine issues, all of which were addressed; it also ran a short independent check at $N=960$ (\cref{app:n960}).

\begin{table}[t]
\centering\small
\caption{Division of labour, as recorded in the project notes.}
\label{tab:roles}
\begin{tabular}{@{}L{3.3cm}L{12.6cm}@{}}
\toprule
Role & Responsibilities \\
\midrule
Authors & choice of problem and sizes; time boxes and stop rules; decisions to continue, stop, or change direction; requirement that every result pass verification by separate code; accepting or rejecting review comments; decisions to release; communication with the CCD maintainers \\
Chat assistant & literature search and reading; experiment design; the search code of every stage (energy kernel, basin hopping, pool and combiner, scheduler and local relaxation) and the verification and release scripts, delivered as packages with instructions for the agent; reading returned data and proposing decisions; drafting reports \\
Agent & deployment and calibration; launching and monitoring long runs; debugging and patches, including the combiner's relabelling fix; verification runs; status reports; executing approved releases \\
Reviewing AI system & review of the plan (nine issues); a short independent check run at $N=960$ \\
\bottomrule
\end{tabular}
\end{table}

\section{Results}
\label{sec:results}

\subsection{Three certified improvements}

\begin{table}[t]
\centering\small
\caption{New configurations and the CCD configurations they improve. Energy: CCD value (rows CCD) or double-precision value with exactly rounded summation (rows new). Certified bound: integer bounds \eqref{eq:cert} for the rationalised configurations (lower bound $L$ for the CCD configurations, upper bound $U$ for the new ones), rounded outwards at the tenth decimal. $\lambda_1$ and $\lambda_{\max}$: smallest and largest Hessian eigenvalues after removing the three rotational modes (whose eigenvalues have magnitude below $10^{-8}$). All six configurations have point group $C_1$ at tolerance $10^{-6}$ and at the CCD's tolerance of $10^{-4}$.}
\label{tab:main}
\setlength{\tabcolsep}{4pt}
\begin{tabular}{@{}rlrrrrrrc@{}}
\toprule
$N$ & & Energy & Certified bound & \makecell[r]{Decrease\\($-\Delta E$)} & $|g|_\infty$ & $\lambda_1$ & $\lambda_{\max}$ & \makecell[c]{5-/7-fold\\points} \\
\midrule
912 & CCD & $400657.6165027$ & $\ge 400657.6165027195$ & & $1.7\times10^{-5}$ & $1.68$ & $6.07\times10^{3}$ & 33 / 21 \\
 & new & $400657.6128841$ & $\le 400657.6128840633$ & $0.0036186$ & $9.7\times10^{-9}$ & $1.91$ & $6.07\times10^{3}$ & 33 / 21 \\
\addlinespace
1382 & CCD & $926576.1030330$ & $\ge 926576.1030329535$ & & $5.2\times10^{-8}$ & $8.23$ & $1.13\times10^{4}$ & 48 / 36 \\
 & new & $926576.0920600$ & $\le 926576.0920599694$ & $0.0109730$ & $3.6\times10^{-9}$ & $8.17$ & $1.13\times10^{4}$ & 48 / 36 \\
\addlinespace
2502 & CCD & $3060842.2403807$ & $\ge 3060842.2403807197$ & & $5.9\times10^{-8}$ & $11.32$ & $2.76\times10^{4}$ & 48 / 36 \\
 & new & $3060842.1789774$ & $\le 3060842.1789773580$ & $0.0614033$ & $9.4\times10^{-9}$ & $11.52$ & $2.76\times10^{4}$ & 48 / 36 \\
\bottomrule
\end{tabular}
\end{table}

\Cref{tab:main} gives the results. For $N=2502$ the certified upper bound for the new configuration, $3060842.1789773580$, is $0.0614033$ below the CCD value and $0.0614034$ below the certified lower bound for the (rationalised) CCD configuration; the corresponding decreases are $0.0109730$ and $0.0109730$ for $N=1382$ and $0.0036186$ and $0.0036187$ for $N=912$, and the relative decreases are $2.01\times10^{-8}$, $1.18\times10^{-8}$ and $9.03\times10^{-9}$. We verified the values three ways: the integer certificate \eqref{eq:cert}; the paper's verification code, which uses exactly rounded summation \citep{shewchuk1997}; and the release verification script, written by the chat assistant during the project and published with the coordinates, run on freshly downloaded CCD files. All agree to $10^{-9}$. The CCD page and coordinate files downloaded on 10 November 2026 are byte-identical to those used during the search on 6 November, and the file for $N=912$ is byte-identical to the one from which its search started.

Numerically, all three new configurations are strict local minima modulo rotations: apart from the three rotational modes (eigenvalues of magnitude below $10^{-10}$), the smallest Hessian eigenvalue is $1.91$ ($N=912$), $8.17$ ($N=1382$) and $11.52$ ($N=2502$). Consistently, adding Gaussian noise of standard deviation $0.05\,\ell$ to every coordinate and re-polishing returned to the same energy (difference $0$ at double precision) for three random seeds. No configuration has a rotation axis of order 2--6, an inversion centre, or a mirror plane.

The improvements do not come from polishing. Polishing the CCD configurations with the same method changes their computed energies by at most one ulp ($0$ for $N=912$, $1.2\times10^{-10}$ for $N=1382$, $4.7\times10^{-10}$ for $N=2502$), that is, not at all within rounding. For scale, the October 2026 update of the CCD lowered the previous best values for $N=1382$ and $N=2502$ by $0.0145$ and $1.2968$ \citep{amore2026efficient,wales2009defect} and reproduced the earlier value for $N=912$ \citep[Table~1]{amore2026efficient}.

\subsection{How the configurations were found}

\begin{figure}[t]
\centering
\includegraphics[width=\linewidth]{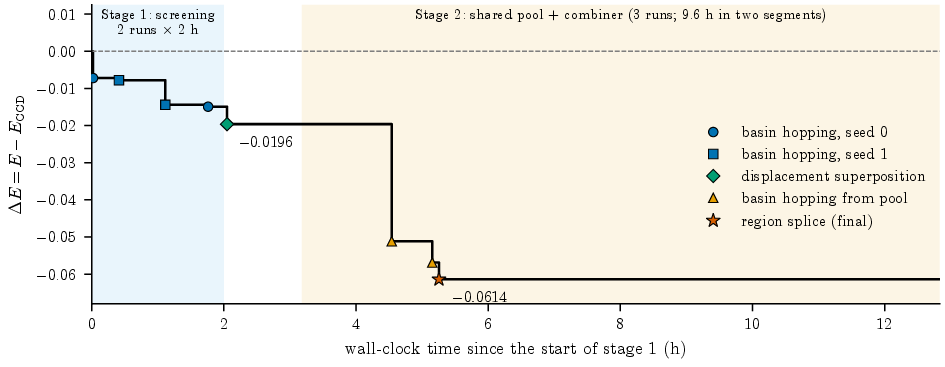}
\caption{Best $\Delta E$ found for $N=2502$ against wall-clock time since the start of stage~1. Markers show the event that produced each new best. The final configuration is a region splice of the stage-2 configuration at $-0.0569$ with a stage-1 configuration at $-0.0096$; it appeared less than three minutes after the combiner was restarted with the corrected relabelling step of \cref{alg:combine} (at $5.2$\,h).}
\label{fig:trajectory}
\end{figure}

\paragraph{$N=912$.} The configuration was found by the probe of smaller sizes, which ran the stage-1 protocol, without pool or combiner, five times on each of four sizes (\cref{app:probe}). One run reached it after $10{,}835$ relaxations ($40$ wall-clock minutes); the other four runs at this size found nothing below the CCD value.

\paragraph{$N=1382$.} The configuration was found directly by basin hopping, $66$ minutes into one of the two stage-1 runs, and only polished afterwards. About $19$ further core-hours in stage~2 ($101{,}646$ relaxations, with the combiner) and $7.5$ core-hours in stage~3 did not improve it.

\paragraph{$N=2502$.} \Cref{fig:trajectory} shows the trajectory. In stage~1 both runs found configurations below the CCD value within $25$ wall-clock minutes (the first after $1.2$ minutes), reaching $-0.0149$ and $-0.0144$ after two hours. Superposing the displacement fields of the two best configurations from one run, $-0.0144$ and $-0.0078$, and relaxing gave $-0.0196$ (the sum of the parents' $\Delta E$ would be $-0.0222$). In stage~2 one run found $-0.0511$ and then $-0.0569$, configurations in which $251$ and $278$ points, respectively, have moved by more than $0.15\,\ell$ from their CCD positions. Label exchanges accumulated during the walk had left the $-0.0569$ configuration with an aligned fraction of only $0.78$, below the combiner's threshold of $0.9$, so the combiner never used it. Once the relabelling step was fixed, the combiner produced the final configuration, $-0.0614$, within three minutes: it spliced into the $-0.0569$ configuration the region in which a stage-1 configuration ($-0.0096$) differs from the CCD configuration, realising about half of that configuration's decrease. After this, the three runs and the combiner found nothing lower in the remaining $7.6$ hours. In total about $12$ core-hours were spent on $N=2502$ before the final configuration appeared ($6$ in stage~1, including a two-hour symmetry-constrained run that contributed nothing, and $6$ in stage~2); including the later stages and the controlled comparison, about $58$ core-hours were spent on this size.

\paragraph{The first step is reproducible; the last one may not be.} Getting below the CCD value for $N=2502$ is easy with full relaxation: in each of the two stage-1 runs and in each of the three two-run arms of the controlled comparison that included full relaxation (\cref{sec:speedup}; the arms shared two random seeds but differed in relaxation mode), a configuration below the CCD value appeared within $25$ wall-clock minutes. Reaching $-0.0614$ is a different matter: it depended on a single stage-2 run, and the other two runs, which differed only in their random seeds, found nothing below $-0.0196$ on their own.

Two observations qualify the role of recombination. First, decreases do not add: none of the six pairwise displacement superpositions tried in stage~1 produced a $\Delta E$ within $2\times10^{-4}$ of the sum of the parents' $\Delta E$, and when the localised changes found in the stage-1 configurations (five locations for each size) were transplanted one at a time into the CCD configuration, three of five lowered the energy for $N=2502$ (but applying them in sequence retained only one) and none of five did for $N=1382$. Second, recombination was not needed to reach the $-0.0196$ configuration: basin hopping from the CCD configuration reached a minimum with the same energy (to $10^{-9}$) in two arms of the controlled comparison that share a random seed and differ only in early termination (ET) of relaxations returning to the starting minimum (\cref{sec:speedup}). Of the $147$ combinations formed in stage~2 after the fix ($114$ for $N=2502$, $33$ for $N=1382$), $10$ were lower than both parents; one of these was the new best, and the other nine reproduced configurations already in the pool (one of them a second copy of the final configuration). The combiner is therefore best viewed as a cheap way to merge progress made in different regions, not as the engine of the search; isolating its contribution would require the ablation discussed in \cref{sec:limitations}.

\subsection{What changed structurally}

\begin{figure}[t]
\centering
\includegraphics[width=\linewidth]{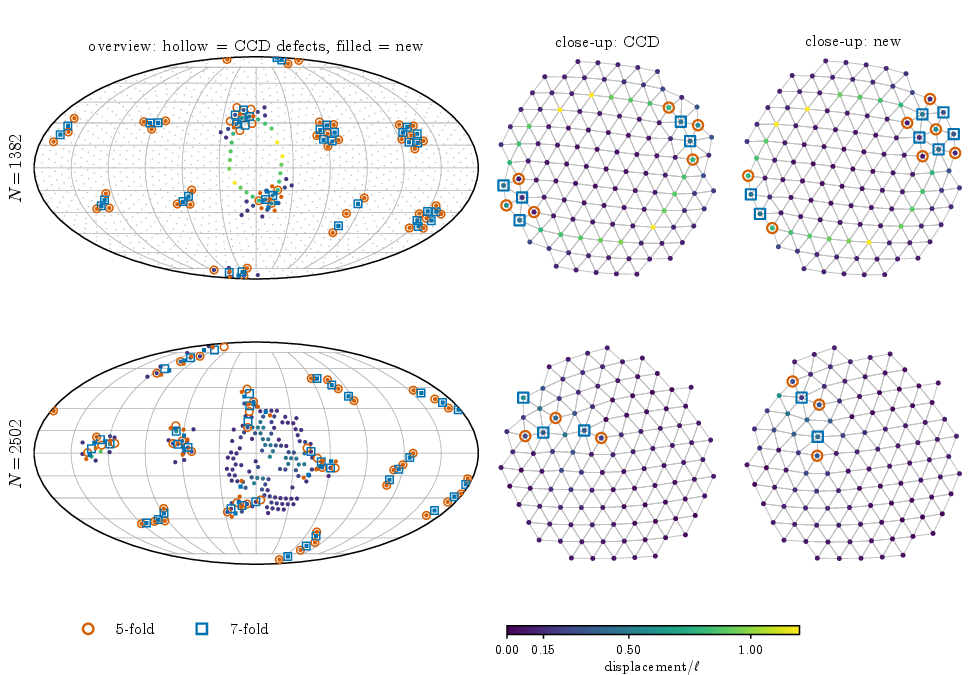}
\caption{Where the new configurations for $N=1382$ and $N=2502$ differ from the CCD configurations. Left: Mollweide projection (centred on the displaced region) of the CCD configuration; points displaced by more than $0.15\,\ell$ in the new configuration are coloured by displacement (in units of $\ell$), hollow markers are defects of the CCD configuration and filled markers those of the new configuration. Middle and right: close-ups of the Delaunay triangulation around the points displaced by more than $0.5\,\ell$, with every point coloured by displacement. Correspondence between the configurations is obtained by optimal rotation and a minimum-cost assignment.}
\label{fig:structure}
\end{figure}

\Cref{fig:structure} shows where the new configurations for $N=1382$ and $N=2502$ differ from the CCD configurations. After optimal rotation and relabelling, $279$ of the $2502$ points move by more than $0.15\,\ell$ and $14$ by more than $0.5\,\ell$ (maximum $0.89\,\ell$); for $N=1382$, $50$ points move by more than $0.15\,\ell$ and $3$ by more than $\ell$ (maximum $1.75\,\ell$), along an arc between two scars, consistent with dislocation glide. The defect inventory is unchanged in all three cases (\cref{tab:main}). For $N=2502$, both configurations consist of twelve scars, each a 5--7--5 chain, and $24$ isolated dislocations; for $N=1382$, both consist of four eleven-defect scars, seven five-defect scars, one isolated 5-fold point, and two isolated dislocations; for $N=912$, both consist of nine five-defect and three three-defect scars. The coordination number of $71$ points ($N=2502$) and $24$ points ($N=1382$) changes, so for these two sizes the improvements are rearrangements of dislocations within the same defect inventory and the same scar compositions. For all three sizes, the CCD configurations were not the lowest minima with their defect inventory and scar compositions. Whether this holds for other entries is open; eight of the eleven sizes searched with full relaxation from the CCD configuration did not improve (\cref{sec:accounting}).

\subsection{Complete accounting of the searches}
\label{sec:accounting}

\Cref{tab:accounting} in \cref{app:sweep} lists every size we searched, with its budget and outcome. Unconstrained basin hopping with full relaxation, started from the CCD configuration, improved two of the seven sizes on which it was run in the project ($N=1382$ and $2502$, but not $2000$, $2192$, $3182$, $3552$, or $4352$). The five sizes that did not improve received $4$ to $10$ core-hours of such search each, against about $31$ ($N=1382$) and $58$ ($N=2502$) core-hours in total for the two that did, so the comparison between them is not balanced. Four further sizes ($5792$, $5872$, $9911$, $9913$) were searched only with the flawed local-relaxation variant described next and remain untested with full relaxation. The total compute of the project, including combiner processes, was about $290$ core-hours; short tests run by the chat assistant in its own sandbox (minutes each) and the reviewing AI system's check (\cref{app:n960}) are not included. The probe of smaller sizes improved one of its four sizes, $N=912$ (\cref{app:probe}); in all, unconstrained basin hopping with full relaxation from the CCD configuration improved three of the eleven sizes on which it was run.

\section{Failures and how they were detected}
\label{sec:failures}

\subsection{A correct speed-up that disabled the search}
\label{sec:speedup}

In the eight-hour second segment of stage~2 the largest size, $N=4352$, managed only $5{,}181$ relaxations, because every move was followed by a relaxation of all $N$ points. For stage~3 the chat assistant proposed local relaxation, predicting relaxations tens of times faster for the largest sizes: after a move, relax only the $80$--$260$ points nearest to it and freeze the rest. The design is sound as arithmetic: frozen--frozen interactions are unchanged, so the energy change can be computed exactly, and the implementation passed its exactness test (the accumulated energy over $40$ consecutive moves matched a full recomputation to $10^{-9}$). We approved its deployment on the strength of this exactness test, without a speed benchmark or a controlled comparison; we count this as a failure of the human--AI workflow, not only of the design. Stage~3 then ran for $79.8$ core-hours over $11$ sizes (about $72$ of them with local relaxation) and found nothing (\cref{tab:sweep}).

\begin{figure}[t]
\centering
\includegraphics[width=0.6\linewidth]{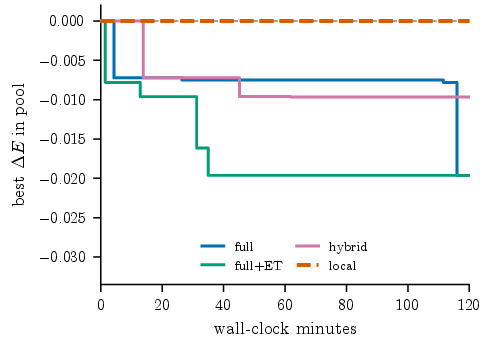}
\caption{Controlled comparison of relaxation modes ($N=2502$, starting from the CCD configuration; four arms, two runs of two hours each, separate pools, no combiner): best $\Delta E$ in each arm's pool against wall-clock minutes. Arms: ``full'', full relaxation; ``full+ET'', full relaxation with ET; ``hybrid'', local relaxation followed by full relaxation when the local result leaves the starting minimum; ``local'', local relaxation. Pass criterion, fixed in advance according to the project notes: a configuration at least $10^{-3}$ below the CCD value within two hours.}
\label{fig:control}
\end{figure}

Because ``nothing'' is also the expected outcome on sizes that are simply hard, the null result alone was uninformative. We therefore ran a controlled comparison on $N=2502$, where full relaxation had reliably found configurations below the CCD value within minutes, and fixed the pass criterion in advance (according to the project notes, in a file that is not independently timestamped): an arm passes if it finds, within two hours, a configuration at least $10^{-3}$ below the CCD value; passing arms are ranked by best $\Delta E$, then by the number of distinct such configurations, then by the time to the first one. The four arms were ``full'' (full relaxation), ``full+ET'' (full relaxation with ET: a relaxation is stopped after 20 iterations if no point has moved by $0.2\,\ell$), ``hybrid'' (local relaxation, followed by full relaxation whenever the local result left the starting minimum, that is, changed the energy by more than $10^{-4}$), and ``local'' (local relaxation); all four used the same two random seeds. \Cref{fig:control,tab:control} show the outcome: the local arm found nothing in four core-hours, while the three arms that included full relaxation passed, full and full+ET both reaching $-0.0196$. Local relaxation was switched off, and by the ranking fixed in advance (six distinct configurations below the CCD value against four) full+ET became the default setting for $N\ge2000$; no further searches were run. Two runs per arm is a small sample; the contrast rests on its size. The two stage-1 runs at $N=2502$ and the three arms that included full relaxation (\cref{sec:results}) each found a configuration below the CCD value after $1.2$ to $27.7$ core-minutes (wall-clock time to the first such configuration times the number of runs); the local arm found none in $240$.

\begin{table}[t]
\centering\small
\caption{Outcome of the controlled comparison (left: wall-clock minutes to the first configuration with $\Delta E\le-10^{-3}$, earlier of the two runs, a dash meaning none within two hours; best $\Delta E$ after two hours) and relaxation of the same 183 random moves applied to the CCD configuration (right: fraction of relaxations that left the starting minimum; mean time per relaxation).}
\label{tab:control}
\begin{tabular}[t]{@{}lrr@{}}
\toprule
Arm & \makecell[r]{first $\Delta E\le-10^{-3}$\\(wall-clock min)} & \makecell[r]{best $\Delta E$\\after 2 h} \\
\midrule
full & 4.3 & $-0.0196$ \\
full+ET & 1.5 & $-0.0196$ \\
hybrid & 13.8 & $-0.0097$ \\
local & -- & $0.0000$ \\
\bottomrule
\end{tabular}
\hspace{2.5em}
\begin{tabular}[t]{@{}lrr@{}}
\toprule
\makecell[l]{Relaxation of the\\same 183 moves} & \makecell[r]{left starting\\minimum (\%)} & \makecell[r]{time per\\relaxation (s)} \\
\midrule
local, $\le260$ points & 3.3 & 1.45 \\
local, 800 points & 7.1 & 17.51 \\
full ($2502$ points) & 7.7 & 2.11 \\
\bottomrule
\end{tabular}
\end{table}

A follow-up measurement points to the mechanism. We applied $183$ random moves to the CCD configuration and relaxed each perturbed configuration three ways (\cref{tab:control}). With $\le260$ active points, only $3.3\%$ of relaxations left the starting minimum (ended at an energy differing from the start by more than $10^{-4}$), against $7.7\%$ for full relaxation; the outcomes are paired, and in $8$ moves only full relaxation left the starting minimum, never the reverse (exact McNemar test, $p=0.008$). Enlarging the active set to $800$ points restored the rate but made relaxation eight times slower than full relaxation; and local relaxation with at most $260$ active points was only $1.45$ times faster than full relaxation ($1.45$\,s against $2.11$\,s at $N=2502$; the predicted speed-up of tens of times was for the largest sizes). This deficit alone is too small to explain the null result: per core-second, local relaxation left the starting minimum at $0.62$ times the rate of full relaxation, which, if the time to a first configuration below the CCD value scales inversely with this rate, would stretch even the slowest of the times above to about $45$ core-minutes, far below the local arm's $240$. The more likely explanation is a bias in the energies. When the local relaxation did leave its starting minimum, its energy exceeded the energy after full relaxation of the same move by $0.1$ to $0.4$, which is $5$ to $60$ times the decreases of $0.007$ to $0.02$ the search was looking for, and by more than $40$ in two relaxations that did not converge. Because both the Metropolis rule and the pool-insertion test used these local energies, they could not see improvements that only appear once the surrounding lattice relaxes; this is consistent with the long-range elastic field of a dislocation, which a frozen boundary cannot accommodate. The hybrid arm, which removes the bias for moves that leave the starting minimum, partly recovered. The general lesson is that the exactness test verified that the code computed what it claimed to compute, and the proxy it optimised (relaxations per hour) improved; neither says anything about whether the search still finds better minima \citep{manheim2018goodhart}.

\subsection{Polishing dividends and spurious records}
\label{sec:dividend}

During stage~3 the pipeline reported configurations below the CCD values for $N=3552$ and $N=5872$. The ease of these results raised suspicion: they appeared without any accepted move. The CCD configurations for these sizes are not converged, and polishing them alone lowers their energies by more than the candidate threshold: they have $|g|_\infty=0.13$ ($N=3552$) and $0.18$ ($N=5872$), and polishing lowers their energies by $1.50\times10^{-5}$ and $1.23\times10^{-5}$ while moving no point by more than $1.2\times10^{-3}\,\ell$. A pipeline that polishes its own candidates but compares them with an unpolished baseline will report this decrease, the polishing dividend, as a discovery. We changed the baseline to the lower of the CCD value and the energy of the polished CCD configuration (the polished baseline), which removed both spurious records; for $N=912$, $1382$ and $2502$, polishing the CCD configuration changes its energy by no more than one ulp. \Cref{fig:polish} in \cref{app:polish} shows the dividend for every CCD entry.

\subsection{Other errors}

\Cref{tab:errors} lists the other errors that occurred. Of the six, five originated in code, a status report or an interpretation from the AI systems, and one (the symmetry hypothesis) came from us. Three were caught by calibration or by anomalies in logged statistics, two by a deliberate test or by the reviewing AI system, and one by checking a report against the run logs; none was found by reading the code. One of them, the combiner's reliance on raw point labels, blocked recombination for two hours and had to be fixed before the final $N=2502$ configuration could be produced.

\begin{table}[t]
\centering\small
\caption{Errors not covered in \cref{sec:speedup,sec:dividend}: origin, how each was detected, and its consequence.}
\label{tab:errors}
\begin{tabular}{@{}L{4.9cm}L{2.0cm}L{4.0cm}L{4.2cm}@{}}
\toprule
Error & Origin & Detected by & Consequence and fix \\
\midrule
Aggressive floating-point compiler optimisation removed the compensation terms of the summation routine (absolute energy error $7.8\times10^{-9}$ at $N=960$) & chat assistant & calibration against the official verifier's scores of three published configurations, before any search & optimisation disabled for that routine only; search kernels unaffected \\
Duplicate detection by rounding energies to $10^{-6}$, below their $5\times10^{-6}$ noise, in a genetic-algorithm variant at $N=960$ & chat assistant & $94\%$ of offspring identical to the start, none accepted & tolerance band of $2\times10^{-5}$; run restarted \\
The stage-2 combiner compared raw point labels, which the walk permutes (aligned fraction $0.78<0.9$), so the best configurations were never combined; the stage-1 superposition script had relabelled points by geometry & chat assistant & combiner tried one pair in two hours & relabelling by geometry (\cref{alg:combine}) \\
The stage-2 status report first stated $78$ candidate pairs and $20$ superpositions, ignoring the combiner's limit of ten configurations & AI systems & checking the report against the run logs & corrected to $54$ pairs and $6$ superpositions \\
Interpretation that the decreases came from independent local rearrangements & AI systems & transplanting each change separately (\cref{sec:results}) & claim withdrawn \\
Hypothesis that ``lower symmetry is better'', inferred from a $T\to D_2\to C_3$ sequence of records at $N=960$ & authors & reviewing AI system: $D_2$ and $C_3$ are not nested & claim withdrawn \\
\bottomrule
\end{tabular}
\end{table}

\section{Discussion}
\label{sec:discussion}

\paragraph{Why this worked here.} For $N=1382$ and $N=2502$, we conjecture that the main reason is the choice of sizes, not algorithmic novelty; our data are consistent with this but do not establish it. The CCD values for the largest sizes were recent; their authors report per-size timings of a few hundred to a few thousand core-seconds in their comparisons with earlier results (the compute behind the CCD values is not stated) and reported that improvements were still appearing \citep{amore2026efficient}. A two-hour screening with a decision rule fixed in advance identified two ``soft'' sizes cheaply, and further compute was concentrated there. The $N=960$ case fits the same pattern: the CCD value there, also from the October 2026 method, fell within a day of being posted as a public challenge, while our ten-hour attempt against the challenge record that replaced it, already lowered by others with the same tools, found nothing. These contrasts differ in more than the compute behind the CCD value, however (size, symmetry, and search code all differ). To remove the differences in symmetry and search code, we ran the stage-1 protocol, with the same code, on the four CCD entries without symmetry just below $N=960$ ($N=900$, $912$, $920$ and $932$): five runs and about $1.1\times10^{5}$ relaxations per size, $4$ to $25$ times the number of stage-1 relaxations per size at the large sizes (\cref{app:probe}). One size improved: after $10{,}835$ relaxations, one run at $N=912$ found the configuration of \cref{sec:results}; the other $19$ runs found nothing below the CCD value, and at each size they found the same lowest distinct minimum (the lowest minimum other than the CCD configuration). According to \citet[Table~1]{amore2026efficient}, the CCD value for $N=912$ predates the October 2026 method, which reproduced it, so an entry that two searches with different methods had reached was still improvable: recency of the CCD value is not necessary for an improvement. At the same time, the probe improved one of four sizes, $N=912$, in only one of its five runs, against two of four sizes in stage~1, each in both of its runs and with fewer relaxations per run. Counts this small do not show that improvements are easier to find at larger $N$, and the two sets of sizes differ in the origin of their CCD values as well as in $N$, so our data do not separate the effect of size from that of the compute behind each CCD value.
The ingredients of the search, basin hopping with replica exchange and crossover-like recombination, are standard \citep{wales1997bh,earl2005,deaven1995,morris1996thomsonga}.

\paragraph{Practices for AI-assisted computational discovery.} Each of the three practices below is tied to an error in this project: calibration caught the summation bug before any search (verification by separate code caught no error; the final claim rests on it); comparison with the polished baseline removed two spurious records after their ease had raised suspicion; and a controlled comparison, run only after stage~3 had spent about $72$ core-hours on local relaxation, diagnosed that change (run before adoption, it would have saved those hours).
\begin{enumerate}
  \item Calibrate the search code, and verify candidates with code separate from the search. Before any search, the energy code had to reproduce the official verifier's scores of three published $N=960$ configurations, which exposed the summation bug (\cref{tab:errors}); for the large-$N$ stages, the search code had to reproduce the CCD values to $10^{-5}$. Every candidate was then recomputed by the release verification script, which shares no code with the search, and the integer certificate lets anyone check the final claim on the released rational coordinates.
  \item Compare candidates with the polished baseline. Otherwise a polishing dividend masquerades as a discovery (\cref{fig:polish}).
  \item Test design changes in controlled comparisons with pass criteria fixed in advance, before deploying them, and record the criteria somewhere timestamped; ours were recorded only in the project notes. A change that is correct, fast and plausible can still disable the search. For $N=2502$, only the controlled comparison distinguished ``this size is hard'' from ``this change broke the search''; for the other sizes searched with the flawed variant the question remains open.
\end{enumerate}
In our experience, the AI systems' speed changed what was cheap: the chat assistant could write a new search variant or verification script, and the agent could deploy it, within a single working session, which made it practical to build controls and checks that we might otherwise have skipped. In this project the scarcest resource seemed to be judgment about which results to trust.

\paragraph{What was specific to working with LLMs.} Several failures, and one weakness of the verification, had a distinctly LLM-assisted character: a regression between code packages delivered at different times (relabelling by geometry in stage~1, raw labels in the stage-2 combiner); a test that checked arithmetic exactness rather than whether the search still worked (\cref{sec:speedup}); an AI-written status report with wrong counts (\cref{tab:errors}); an unverified speed prediction (``tens of times faster'' for the largest sizes, measured $1.45$ times at $N=2502$); and verification code written by the same model family as the search (\cref{sec:limitations}). The reviewing AI system, in turn, caught a reasoning error of ours (\cref{tab:errors}), which suggests, as a hypothesis, that a different AI system should critique the plan and the conclusions. Agent benchmarks score outcomes against fixed graders \citep{chan2024mlebench,wijk2024rebench,starace2025paperbench,chen2024scienceagentbench}; open-ended discovery has no grader, and failures of the kinds above are not exposed by an outcome grader.

\paragraph{Open questions.} Whether recombination helps at all when decreases are so far from additive, and if so why, is open: one splice supplied the last $7\%$ of the $N=2502$ decrease, but we ran no ablation; elastic interactions between rearranged regions are a natural starting point. It is also open whether the many CCD sizes on which we found nothing are close to optimal or were simply under-searched by our budgets.

\subsection{Limitations}
\label{sec:limitations}

\begin{itemize}
  \item We ran no equal-budget comparison. We did not run the October 2026 method of \citet{amore2026efficient}, or plain basin hopping, with the same budget on the same sizes, so we cannot say whether our pipeline is more efficient than theirs; our budget on $N=2502$ was larger than the per-size timings they report.
  \item There were few independent runs. The success at $N=2502$ depended on one run that found the $-0.0511$ and $-0.0569$ configurations; runs differing only in the random seed did not. The success at $N=912$ came from one of five runs. The controlled comparison used two runs per arm, with the same two seeds in every arm, on a laptop whose performance and efficiency cores run at different speeds.
  \item We ran no ablations. We did not separately remove the pool, the combiner, or replica exchange, so we cannot attribute the decrease to individual components.
  \item The success rate is low and the effect small. Three of the eleven sizes searched with full relaxation from the CCD configuration improved; the relative decreases are of order $10^{-8}$.
  \item The CCD maintainers have not yet confirmed or posted the configurations; ongoing searches by others may since have found lower energies.
  \item The release verification script was written by the chat assistant, and the paper's verification code and the certification script with Claude Code. They share no code with the search, and the certificate is short integer arithmetic that can be audited by hand; we release the rational coordinates it uses, but the scripts are not independent of the model.
  \item The evidence on AI-assisted research comes from one project on one problem, and our account of the AI systems' contribution is qualitative: we did not measure how much time the AI systems saved or what fraction of the chat assistant's proposals were adopted, and we cannot say what we would have achieved on the same budget without them, or with standard tools \citep{gmin}.
\end{itemize}

\section{Conclusion}

We reported configurations for the Thomson problem with $N=912$, $1382$ and $2502$ whose energies are certifiably below the CCD values, found on a laptop by a search pipeline that the AI systems wrote and ran under our direction; its first stage alone, replica-exchange basin hopping without pool or recombination, found the configurations for $N=912$ (in one of five runs) and $N=1382$. The AI systems made the search fast to build and easy to extend; what made its output a claim others can trust were the three practices of \cref{sec:discussion}: calibration and verification by code separate from the search, ending in an integer certificate that anyone can re-run on the released rational coordinates; comparison with the polished baseline; and controlled comparisons with pass criteria fixed in advance (the one we ran showed that a correct speed-up had disabled the search).

\subsubsection*{Broader Impact Statement}
This work concerns a mathematical optimisation benchmark and poses no direct risk of harm. Its indirect relevance is to how AI-assisted results are reported. Fast generation of plausible code and plausible results makes it easier to publish claims that are wrong; the verification practices and complete reporting used here are intended to counter that, and we encourage similar disclosure of failed attempts and of the human--AI division of labour in other AI-assisted work.

\subsubsection*{Use of AI systems}
The search, analysis and visualisation code used during the project was written by Claude (Anthropic) through the claude.ai chat interface and deployed, run and patched by the Claude Code agent, under our direction (\cref{tab:roles}); the reviewing AI system, used separately, reviewed the plan and ran a short independent check. The project notes do not record the model versions or the identity of the reviewing AI system. The paper's verification code, the other analysis code written for this paper, and a first draft of this manuscript were produced with Claude Opus 5.5 through Claude Code, which also proposed and ran the probe of smaller sizes with the project's search code (\cref{app:probe}); all of it was checked and revised by the authors, who take full responsibility for the content.

\subsubsection*{Data and code availability}
Coordinates of the $N=1382$ and $N=2502$ configurations, the release verification script, which shares no code with the search, and the release history are public at \repo{} and archived on Zenodo (\zenodo\ifpreprint; \citealt{wang2026zenodo}\fi). The search code, run logs, pool contents, controlled-comparison data, the coordinates and certificate of the $N=912$ configuration, and the analysis scripts of this paper are provided as supplementary material (\cref{app:repro}).

\ifpreprint
\subsubsection*{Acknowledgments}
We thank David J.\ Wales and colleagues for maintaining the CCD and making the coordinates of every entry public, which made this work possible.
\fi

\bibliography{references}

\begin{thebibliography}{70}
\providecommand{\natexlab}[1]{#1}
\providecommand{\url}[1]{\texttt{#1}}
\expandafter\ifx\csname urlstyle\endcsname\relax
  \providecommand{\doi}[1]{doi: #1}\else
  \providecommand{\doi}{doi: \begingroup \urlstyle{rm}\Url}\fi

\bibitem[{Abreto}(2026)]{abreto2026p007}
{Abreto}.
\newblock Few {S}hots p007 entry \texttt{thomson960-d2-ga}: a {$D_2$}-symmetric
  configuration of 960 points, verifier score 444368.4040805305.
\newblock \url{https://github.com/few-shots/few-shots/pull/15}, 2026.
\newblock Verified 27 October 2026, 20:50 UTC; agent Claude Code. Accessed 10
  November 2026.

\bibitem[Absil et~al.(2008)Absil, Mahony, and Sepulchre]{absil2008}
P.-A. Absil, R.~Mahony, and R.~Sepulchre.
\newblock \emph{Optimization Algorithms on Matrix Manifolds}.
\newblock Princeton University Press, Princeton, NJ, 2008.

\bibitem[Altschuler et~al.(1997)Altschuler, Williams, Ratner, Tipton, Stong,
  Dowla, and Wooten]{altschuler1997}
E.~L. Altschuler, T.~J. Williams, E.~R. Ratner, R.~Tipton, R.~Stong, F.~Dowla,
  and F.~Wooten.
\newblock Possible global minimum lattice configurations for {T}homson's
  problem of charges on a sphere.
\newblock \emph{Physical Review Letters}, 78\penalty0 (14):\penalty0
  2681--2685, 1997.

\bibitem[Altschuler \& P{\'e}rez-Garrido(2006)Altschuler and
  P{\'e}rez-Garrido]{altschuler2006}
Eric~Lewin Altschuler and Antonio P{\'e}rez-Garrido.
\newblock Defect-free global minima in {T}homson's problem of charges on a
  sphere.
\newblock \emph{Physical Review E}, 73\penalty0 (3):\penalty0 036108, 2006.

\bibitem[Amore \& Wales(2026)Amore and Wales]{amore2026efficient}
Paolo Amore and David~J. Wales.
\newblock Efficient searches for low-energy structures in clusters with
  thousands of particles: Application to the {T}homson problem.
\newblock \emph{The Journal of Chemical Physics}, 165\penalty0 (16):\penalty0
  164104, 2026.
\newblock Also arXiv:2610.08122.

\bibitem[Amore et~al.(2025)Amore, Figueroa, Diaz, L{\'o}pez, and
  Vincent]{amore2025landscape}
Paolo Amore, Victor Figueroa, Enrique Diaz, Jorge~A. L{\'o}pez, and Trevor
  Vincent.
\newblock Exploring the energy landscape of the {T}homson problem: local minima
  and stationary states.
\newblock arXiv preprint arXiv:2506.08398, 2025.

\bibitem[{Anthropic}(2025)]{anthropic2025claudecode}
{Anthropic}.
\newblock Claude {C}ode.
\newblock \url{https://www.anthropic.com/claude-code}, 2025.
\newblock Agentic coding tool; accessed November 2026.

\bibitem[Barzilai \& Borwein(1988)Barzilai and Borwein]{barzilai1988}
Jonathan Barzilai and Jonathan~M. Borwein.
\newblock Two-point step size gradient methods.
\newblock \emph{IMA Journal of Numerical Analysis}, 8\penalty0 (1):\penalty0
  141--148, 1988.

\bibitem[Bausch et~al.(2003)Bausch, Bowick, Cacciuto, Dinsmore, Hsu, Nelson,
  Nikolaides, Travesset, and Weitz]{bausch2003}
A.~R. Bausch, M.~J. Bowick, A.~Cacciuto, A.~D. Dinsmore, M.~F. Hsu, D.~R.
  Nelson, M.~G. Nikolaides, A.~Travesset, and D.~A. Weitz.
\newblock Grain boundary scars and spherical crystallography.
\newblock \emph{Science}, 299\penalty0 (5613):\penalty0 1716--1718, 2003.

\bibitem[Bianchi et~al.(2026)Bianchi, Kwon, Pappu, and
  Zou]{bianchi2026harnessing}
Federico Bianchi, Yongchan Kwon, Aneesh Pappu, and James Zou.
\newblock Harnessing the collective intelligence of {AI} agents in the wild for
  new discoveries.
\newblock arXiv preprint arXiv:2606.10402, 2026.

\bibitem[Borodachov et~al.(2019)Borodachov, Hardin, and Saff]{borodachov2019}
Sergiy~V. Borodachov, Douglas~P. Hardin, and Edward~B. Saff.
\newblock \emph{Discrete Energy on Rectifiable Sets}.
\newblock Springer Monographs in Mathematics. Springer, New York, 2019.

\bibitem[Bowick et~al.(2002)Bowick, Cacciuto, Nelson, and
  Travesset]{bowick2002}
Mark Bowick, Angelo Cacciuto, David~R. Nelson, and Alex Travesset.
\newblock Crystalline order on a sphere and the generalized {T}homson problem.
\newblock \emph{Physical Review Letters}, 89\penalty0 (18):\penalty0 185502,
  2002.

\bibitem[Bowick et~al.(2000)Bowick, Nelson, and Travesset]{bowick2000}
Mark~J. Bowick, David~R. Nelson, and Alex Travesset.
\newblock Interacting topological defects on frozen topographies.
\newblock \emph{Physical Review B}, 62\penalty0 (13):\penalty0 8738--8751,
  2000.

\bibitem[Byrd et~al.(1995)Byrd, Lu, Nocedal, and Zhu]{byrd1995}
Richard~H. Byrd, Peihuang Lu, Jorge Nocedal, and Ciyou Zhu.
\newblock A limited memory algorithm for bound constrained optimization.
\newblock \emph{SIAM Journal on Scientific Computing}, 16\penalty0
  (5):\penalty0 1190--1208, 1995.

\bibitem[Chan et~al.(2025)Chan, Chowdhury, Jaffe, Aung, Sherburn, Mays,
  Starace, Liu, Maksin, Patwardhan, Weng, and M{\k{a}}dry]{chan2024mlebench}
Jun~Shern Chan, Neil Chowdhury, Oliver Jaffe, James Aung, Dane Sherburn, Evan
  Mays, Giulio Starace, Kevin Liu, Leon Maksin, Tejal Patwardhan, Lilian Weng,
  and Aleksander M{\k{a}}dry.
\newblock {MLE-bench}: Evaluating machine learning agents on machine learning
  engineering.
\newblock In \emph{The Thirteenth International Conference on Learning
  Representations (ICLR)}, 2025.

\bibitem[Charton et~al.(2024)Charton, Ellenberg, Wagner, and
  Williamson]{charton2024patternboost}
Fran{\c{c}}ois Charton, Jordan~S. Ellenberg, Adam~Zsolt Wagner, and Geordie
  Williamson.
\newblock {PatternBoost}: Constructions in mathematics with a little help from
  {AI}.
\newblock arXiv preprint arXiv:2411.00566, 2024.

\bibitem[Chen et~al.(2025)Chen, Chen, Ning, Zhang, Wang, Yu, Li, Liao, Wei, Lu,
  Dey, Xue, Baker, Burns, Adu-Ampratwum, Huang, Ning, Gao, Su, and
  Sun]{chen2024scienceagentbench}
Ziru Chen, Shijie Chen, Yuting Ning, Qianheng Zhang, Boshi Wang, Botao Yu,
  Yifei Li, Zeyi Liao, Chen Wei, Zitong Lu, Vishal Dey, Mingyi Xue, Frazier~N.
  Baker, Benjamin Burns, Daniel Adu-Ampratwum, Xuhui Huang, Xia Ning, Song Gao,
  Yu~Su, and Huan Sun.
\newblock {ScienceAgentBench}: Toward rigorous assessment of language agents
  for data-driven scientific discovery.
\newblock In \emph{The Thirteenth International Conference on Learning
  Representations (ICLR)}, 2025.

\bibitem[Cohn \& Kumar(2007)Cohn and Kumar]{cohn2007}
Henry Cohn and Abhinav Kumar.
\newblock Universally optimal distribution of points on spheres.
\newblock \emph{Journal of the American Mathematical Society}, 20\penalty0
  (1):\penalty0 99--148, 2007.

\bibitem[Davies et~al.(2021)Davies, Veli{\v{c}}kovi{\'c}, Buesing, Blackwell,
  Zheng, Toma{\v{s}}ev, Tanburn, Battaglia, Blundell, Juh{\'a}sz, Lackenby,
  Williamson, Hassabis, and Kohli]{davies2021}
Alex Davies, Petar Veli{\v{c}}kovi{\'c}, Lars Buesing, Sam Blackwell, Daniel
  Zheng, Nenad Toma{\v{s}}ev, Richard Tanburn, Peter Battaglia, Charles
  Blundell, Andr{\'a}s Juh{\'a}sz, Marc Lackenby, Geordie Williamson, Demis
  Hassabis, and Pushmeet Kohli.
\newblock Advancing mathematics by guiding human intuition with {AI}.
\newblock \emph{Nature}, 600:\penalty0 70--74, 2021.

\bibitem[Deaven \& Ho(1995)Deaven and Ho]{deaven1995}
D.~M. Deaven and K.~M. Ho.
\newblock Molecular geometry optimization with a genetic algorithm.
\newblock \emph{Physical Review Letters}, 75\penalty0 (2):\penalty0 288--291,
  1995.

\bibitem[Doye et~al.(1999)Doye, Miller, and Wales]{doye1999lj38}
Jonathan P.~K. Doye, Mark~A. Miller, and David~J. Wales.
\newblock The double-funnel energy landscape of the 38-atom {L}ennard-{J}ones
  cluster.
\newblock \emph{The Journal of Chemical Physics}, 110\penalty0 (14):\penalty0
  6896--6906, 1999.

\bibitem[Earl \& Deem(2005)Earl and Deem]{earl2005}
David~J. Earl and Michael~W. Deem.
\newblock Parallel tempering: Theory, applications, and new perspectives.
\newblock \emph{Physical Chemistry Chemical Physics}, 7\penalty0 (23):\penalty0
  3910--3916, 2005.

\bibitem[Erber \& Hockney(1991)Erber and Hockney]{erber1991}
T.~Erber and G.~M. Hockney.
\newblock Equilibrium configurations of {$N$} equal charges on a sphere.
\newblock \emph{Journal of Physics A: Mathematical and General}, 24\penalty0
  (23):\penalty0 L1369--L1377, 1991.

\bibitem[Fawzi et~al.(2022)Fawzi, Balog, Huang, Hubert, Romera-Paredes,
  Barekatain, Novikov, Ruiz, Schrittwieser, Swirszcz, Silver, Hassabis, and
  Kohli]{fawzi2022alphatensor}
Alhussein Fawzi, Matej Balog, Aja Huang, Thomas Hubert, Bernardino
  Romera-Paredes, Mohammadamin Barekatain, Alexander Novikov, Francisco J.~R.
  Ruiz, Julian Schrittwieser, Grzegorz Swirszcz, David Silver, Demis Hassabis,
  and Pushmeet Kohli.
\newblock Discovering faster matrix multiplication algorithms with
  reinforcement learning.
\newblock \emph{Nature}, 610:\penalty0 47--53, 2022.

\bibitem[{Few Shots}(2026)]{fewshots2026}
{Few Shots}.
\newblock Problem p007: {T}homson problem for 960 points.
\newblock
  \url{https://github.com/few-shots/few-shots/tree/main/problems/p007-thomson-problem},
  2026.
\newblock Public AI research challenge; problem opened 27 October 2026,
  integer-arithmetic verifier, baseline the CCD configuration. Accessed 10
  November 2026.

\bibitem[Georgiev et~al.(2025)Georgiev, G{\'o}mez-Serrano, Tao, and
  Wagner]{georgiev2025}
Bogdan Georgiev, Javier G{\'o}mez-Serrano, Terence Tao, and Adam~Zsolt Wagner.
\newblock Mathematical exploration and discovery at scale.
\newblock arXiv preprint arXiv:2511.02864, 2025.

\bibitem[Gottweis et~al.(2025)Gottweis, Weng, Daryin, Tu, Sirkovic,
  et~al.]{gottweis2025coscientist}
Juraj Gottweis, Wei-Hung Weng, Alexander Daryin, Tao Tu, Petar Sirkovic, et~al.
\newblock Accelerating scientific discovery with {Co-Scientist}.
\newblock arXiv preprint arXiv:2502.18864, 2025.

\bibitem[Griffiths et~al.(2017)Griffiths, Niblett, and
  Wales]{griffiths2017alignment}
Matthew Griffiths, Samuel~P. Niblett, and David~J. Wales.
\newblock Optimal alignment of structures for finite and periodic systems.
\newblock \emph{Journal of Chemical Theory and Computation}, 13\penalty0
  (10):\penalty0 4914--4931, 2017.

\bibitem[Harris et~al.(2020)Harris, Millman, van~der Walt,
  et~al.]{harris2020numpy}
Charles~R. Harris, K.~Jarrod Millman, St{\'e}fan~J. van~der Walt, et~al.
\newblock Array programming with {NumPy}.
\newblock \emph{Nature}, 585:\penalty0 357--362, 2020.

\bibitem[Henderson et~al.(2018)Henderson, Islam, Bachman, Pineau, Precup, and
  Meger]{henderson2018}
Peter Henderson, Riashat Islam, Philip Bachman, Joelle Pineau, Doina Precup,
  and David Meger.
\newblock Deep reinforcement learning that matters.
\newblock In \emph{Proceedings of the AAAI Conference on Artificial
  Intelligence}, volume~32, 2018.

\bibitem[Hukushima \& Nemoto(1996)Hukushima and Nemoto]{hukushima1996}
Koji Hukushima and Koji Nemoto.
\newblock Exchange {M}onte {C}arlo method and application to spin glass
  simulations.
\newblock \emph{Journal of the Physical Society of Japan}, 65\penalty0
  (6):\penalty0 1604--1608, 1996.

\bibitem[Hunter(2007)]{hunter2007matplotlib}
John~D. Hunter.
\newblock Matplotlib: A {2D} graphics environment.
\newblock \emph{Computing in Science \& Engineering}, 9\penalty0 (3):\penalty0
  90--95, 2007.

\bibitem[Jimenez et~al.(2024)Jimenez, Yang, Wettig, Yao, Pei, Press, and
  Narasimhan]{jimenez2024swebench}
Carlos~E. Jimenez, John Yang, Alexander Wettig, Shunyu Yao, Kexin Pei, Ofir
  Press, and Karthik Narasimhan.
\newblock {SWE}-bench: Can language models resolve real-world {GitHub} issues?
\newblock In \emph{International Conference on Learning Representations
  (ICLR)}, 2024.

\bibitem[Kabsch(1976)]{kabsch1976}
W.~Kabsch.
\newblock A solution for the best rotation to relate two sets of vectors.
\newblock \emph{Acta Crystallographica Section A}, 32\penalty0 (5):\penalty0
  922--923, 1976.

\bibitem[Kahan(1965)]{kahan1965}
William Kahan.
\newblock Further remarks on reducing truncation errors.
\newblock \emph{Communications of the ACM}, 8\penalty0 (1):\penalty0 40, 1965.

\bibitem[Kryvonos et~al.(2026)Kryvonos, Liehr, and Taylor]{kryvonos2026eight}
Liudmyla Kryvonos, Lukas Liehr, and Mitchell~A. Taylor.
\newblock Energy minimization for eight points on the sphere.
\newblock arXiv preprint arXiv:2609.22077, 2026.

\bibitem[Kuhn(1955)]{kuhn1955}
H.~W. Kuhn.
\newblock The {H}ungarian method for the assignment problem.
\newblock \emph{Naval Research Logistics Quarterly}, 2\penalty0
  (1--2):\penalty0 83--97, 1955.

\bibitem[Li \& Scheraga(1987)Li and Scheraga]{li1987}
Zhenqin Li and Harold~A. Scheraga.
\newblock {M}onte {C}arlo-minimization approach to the multiple-minima problem
  in protein folding.
\newblock \emph{Proceedings of the National Academy of Sciences}, 84\penalty0
  (19):\penalty0 6611--6615, 1987.

\bibitem[Liu \& Nocedal(1989)Liu and Nocedal]{liu1989}
Dong~C. Liu and Jorge Nocedal.
\newblock On the limited memory {BFGS} method for large scale optimization.
\newblock \emph{Mathematical Programming}, 45\penalty0 (1--3):\penalty0
  503--528, 1989.

\bibitem[Lu et~al.(2024)Lu, Lu, Lange, Foerster, Clune, and
  Ha]{lu2024aiscientist}
Chris Lu, Cong Lu, Robert~Tjarko Lange, Jakob Foerster, Jeff Clune, and David
  Ha.
\newblock The {AI} {S}cientist: Towards fully automated open-ended scientific
  discovery.
\newblock arXiv preprint arXiv:2408.06292, 2024.

\bibitem[Manheim \& Garrabrant(2018)Manheim and
  Garrabrant]{manheim2018goodhart}
David Manheim and Scott Garrabrant.
\newblock Categorizing variants of {G}oodhart's law.
\newblock arXiv preprint arXiv:1803.04585, 2018.

\bibitem[Morris et~al.(1996)Morris, Deaven, and Ho]{morris1996thomsonga}
J.~R. Morris, D.~M. Deaven, and K.~M. Ho.
\newblock Genetic-algorithm energy minimization for point charges on a sphere.
\newblock \emph{Physical Review B}, 53\penalty0 (4):\penalty0 R1740--R1743,
  1996.

\bibitem[Neumaier(1974)]{neumaier1974}
Arnold Neumaier.
\newblock Rundungsfehleranalyse einiger {V}erfahren zur {S}ummation endlicher
  {S}ummen.
\newblock \emph{ZAMM -- Zeitschrift f{\"u}r Angewandte Mathematik und
  Mechanik}, 54\penalty0 (1):\penalty0 39--51, 1974.

\bibitem[Nosek et~al.(2018)Nosek, Ebersole, DeHaven, and
  Mellor]{nosek2018prereg}
Brian~A. Nosek, Charles~R. Ebersole, Alexander~C. DeHaven, and David~T. Mellor.
\newblock The preregistration revolution.
\newblock \emph{Proceedings of the National Academy of Sciences}, 115\penalty0
  (11):\penalty0 2600--2606, 2018.

\bibitem[Novikov et~al.(2025)Novikov, V{\~u}, Eisenberger, Dupont, Huang,
  Wagner, Shirobokov, Kozlovskii, Ruiz, Mehrabian, Kumar, See, Chaudhuri,
  Holland, Davies, Nowozin, Kohli, and Balog]{novikov2025alphaevolve}
Alexander Novikov, Ng{\^a}n V{\~u}, Marvin Eisenberger, Emilien Dupont, Po-Sen
  Huang, Adam~Zsolt Wagner, Sergey Shirobokov, Borislav Kozlovskii, Francisco
  J.~R. Ruiz, Abbas Mehrabian, M.~Pawan Kumar, Abigail See, Swarat Chaudhuri,
  George Holland, Alex Davies, Sebastian Nowozin, Pushmeet Kohli, and Matej
  Balog.
\newblock {AlphaEvolve}: A coding agent for scientific and algorithmic
  discovery.
\newblock arXiv preprint arXiv:2506.13131, 2025.

\bibitem[P{\'e}rez-Garrido \& Moore(1999)P{\'e}rez-Garrido and
  Moore]{perezgarrido1999}
A.~P{\'e}rez-Garrido and M.~A. Moore.
\newblock Symmetric patterns of dislocations in {T}homson's problem.
\newblock \emph{Physical Review B}, 60\penalty0 (23):\penalty0 15628--15631,
  1999.

\bibitem[P{\'e}rez-Garrido et~al.(1997)P{\'e}rez-Garrido, Dodgson, and
  Moore]{perezgarrido1997}
A.~P{\'e}rez-Garrido, M.~J.~W. Dodgson, and M.~A. Moore.
\newblock Influence of dislocations in {T}homson's problem.
\newblock \emph{Physical Review B}, 56\penalty0 (7):\penalty0 3640--3643, 1997.

\bibitem[Romera-Paredes et~al.(2024)Romera-Paredes, Barekatain, Novikov, Balog,
  Kumar, Dupont, Ruiz, Ellenberg, Wang, Fawzi, Kohli, and
  Fawzi]{romera2024funsearch}
Bernardino Romera-Paredes, Mohammadamin Barekatain, Alexander Novikov, Matej
  Balog, M.~Pawan Kumar, Emilien Dupont, Francisco J.~R. Ruiz, Jordan~S.
  Ellenberg, Pengming Wang, Omar Fawzi, Pushmeet Kohli, and Alhussein Fawzi.
\newblock Mathematical discoveries from program search with large language
  models.
\newblock \emph{Nature}, 625:\penalty0 468--475, 2024.

\bibitem[{Ronnie2025}(2026)]{ronnie2025p007}
{Ronnie2025}.
\newblock Few {S}hots p007 entry \texttt{c3-replica-exchange-scars}: a
  {$C_3$}-symmetric configuration of 960 points, verifier score
  444368.3924061973.
\newblock \url{https://github.com/few-shots/few-shots/pull/17}, 2026.
\newblock Verified 28 October 2026, 00:35 UTC; agent Claude Code, with several
  declared models. Accessed 10 November 2026.

\bibitem[Roshal et~al.(2015)Roshal, Myasnikova, and Rochal]{roshal2015}
D.~S. Roshal, A.~E. Myasnikova, and S.~B. Rochal.
\newblock Slightly broken icosahedral symmetry advances {T}homson problem.
\newblock \emph{Physics Letters A}, 379:\penalty0 372--376, 2015.

\bibitem[Saff \& Kuijlaars(1997)Saff and Kuijlaars]{saff1997}
Edward~B. Saff and Arno B.~J. Kuijlaars.
\newblock Distributing many points on a sphere.
\newblock \emph{The Mathematical Intelligencer}, 19\penalty0 (1):\penalty0
  5--11, 1997.

\bibitem[Schwartz(2013)]{schwartz2013}
Richard~Evan Schwartz.
\newblock The five-electron case of {T}homson's problem.
\newblock \emph{Experimental Mathematics}, 22\penalty0 (2):\penalty0 157--186,
  2013.

\bibitem[Shewchuk(1997)]{shewchuk1997}
Jonathan~Richard Shewchuk.
\newblock Adaptive precision floating-point arithmetic and fast robust
  geometric predicates.
\newblock \emph{Discrete \& Computational Geometry}, 18\penalty0 (3):\penalty0
  305--363, 1997.

\bibitem[Smale(1998)]{smale1998}
Steve Smale.
\newblock Mathematical problems for the next century.
\newblock \emph{The Mathematical Intelligencer}, 20\penalty0 (2):\penalty0
  7--15, 1998.

\bibitem[Starace et~al.(2025)Starace, Jaffe, Sherburn, Aung, Chan, Maksin,
  Dias, Mays, Kinsella, Thompson, Heidecke, Glaese, and
  Patwardhan]{starace2025paperbench}
Giulio Starace, Oliver Jaffe, Dane Sherburn, James Aung, Jun~Shern Chan, Leon
  Maksin, Rachel Dias, Evan Mays, Benjamin Kinsella, Wyatt Thompson, Johannes
  Heidecke, Amelia Glaese, and Tejal Patwardhan.
\newblock {PaperBench}: Evaluating {AI}'s ability to replicate {AI} research.
\newblock In \emph{Proceedings of the 42nd International Conference on Machine
  Learning}, volume 267 of \emph{Proceedings of Machine Learning Research},
  pp.\  56843--56873. PMLR, 2025.

\bibitem[Swendsen \& Wang(1986)Swendsen and Wang]{swendsen1986}
Robert~H. Swendsen and Jian-Sheng Wang.
\newblock Replica {M}onte {C}arlo simulation of spin-glasses.
\newblock \emph{Physical Review Letters}, 57\penalty0 (21):\penalty0
  2607--2609, 1986.

\bibitem[Thomson(1904)]{thomson1904}
J.~J. Thomson.
\newblock On the structure of the atom: an investigation of the stability and
  periods of oscillation of a number of corpuscles arranged at equal intervals
  around the circumference of a circle; with application of the results to the
  theory of atomic structure.
\newblock \emph{The London, Edinburgh, and Dublin Philosophical Magazine and
  Journal of Science}, 7\penalty0 (39):\penalty0 237--265, 1904.

\bibitem[Tooby-Smith \& Zughaid(2026)Tooby-Smith and
  Zughaid]{toobysmith2026warrant}
Joseph Tooby-Smith and Alex Zughaid.
\newblock A warrant for the {$N = 8$} {T}homson problem, in {L}ean 4.
\newblock \url{https://github.com/jstoobysmith/Thomson-N-8-Warrant}, 2026.
\newblock Accessed 10 November 2026.

\bibitem[Tran(2026)]{tran2026lean}
Hung Tran.
\newblock A {L}ean proof of the {T}homson problem for seven electrons.
\newblock Vals AI blog, \url{https://www.vals.ai/blogs/thomson-n7-lean-proof},
  2026.
\newblock Posted 2026-09-28.

\bibitem[Virtanen et~al.(2020)Virtanen, Gommers, Oliphant,
  et~al.]{virtanen2020scipy}
Pauli Virtanen, Ralf Gommers, Travis~E. Oliphant, et~al.
\newblock {SciPy} 1.0: fundamental algorithms for scientific computing in
  {P}ython.
\newblock \emph{Nature Methods}, 17:\penalty0 261--272, 2020.

\bibitem[Wagner(2021)]{wagner2021}
Adam~Zsolt Wagner.
\newblock Constructions in combinatorics via neural networks.
\newblock arXiv preprint arXiv:2104.14516, 2021.

\bibitem[Wales(2003)]{wales2003book}
David~J. Wales.
\newblock \emph{Energy Landscapes: Applications to Clusters, Biomolecules and
  Glasses}.
\newblock Cambridge University Press, Cambridge, 2003.

\bibitem[Wales(2026)]{gmin}
David~J. Wales.
\newblock {GMIN}: A program for finding global minima and calculating
  thermodynamic properties from basin-sampling.
\newblock \url{https://www-wales.ch.cam.ac.uk/GMIN/}, 2026.
\newblock Accessed 10 November 2026.

\bibitem[Wales \& Doye(1997)Wales and Doye]{wales1997bh}
David~J. Wales and Jonathan P.~K. Doye.
\newblock Global optimization by basin-hopping and the lowest energy structures
  of {L}ennard-{J}ones clusters containing up to 110 atoms.
\newblock \emph{The Journal of Physical Chemistry A}, 101\penalty0
  (28):\penalty0 5111--5116, 1997.

\bibitem[Wales \& Ulker(2006)Wales and Ulker]{wales2006thomson}
David~J. Wales and Sidika Ulker.
\newblock Structure and dynamics of spherical crystals characterized for the
  {T}homson problem.
\newblock \emph{Physical Review B}, 74\penalty0 (21):\penalty0 212101, 2006.

\bibitem[Wales et~al.(2009)Wales, McKay, and Altschuler]{wales2009defect}
David~J. Wales, Hayley McKay, and Eric~L. Altschuler.
\newblock Defect motifs for spherical topologies.
\newblock \emph{Physical Review B}, 79\penalty0 (22):\penalty0 224115, 2009.

\bibitem[Wales et~al.(2026)]{ccd}
David~J. Wales et~al.
\newblock The {C}ambridge {C}luster {D}atabase: Riesz energy database, $s=1$
  ({T}homson problem).
\newblock
  \url{https://www-wales.ch.cam.ac.uk/~wales/CCD/Thomson/site/s_1/index.html},
  2026.
\newblock Accessed 2026-11-06 and 2026-11-10 (identical page content).

\bibitem[Wang(2026)]{wang2026zenodo}
Tao Wang.
\newblock Lower-energy configurations for the {T}homson problem: {$N = 2502$}
  and {$N = 1382$}.
\newblock Zenodo dataset (concept DOI 10.5281/zenodo.23996279), version 1.1.0:
  DOI 10.5281/zenodo.24014200; code and coordinates at
  \url{https://github.com/jesse87wang-bit/thomson-records}, 2026.

\bibitem[Wijk et~al.(2025)Wijk, Lin, Becker, Jawhar, Parikh, Broadley, Chan,
  Chen, Clymer, Dhyani, Ericheva, Garcia, Goodrich, Jurkovic, Kinniment, Lajko,
  Nix, Sato, Saunders, Taran, West, and Barnes]{wijk2024rebench}
Hjalmar Wijk, Tao Lin, Joel Becker, Sami Jawhar, Neev Parikh, Thomas Broadley,
  Lawrence Chan, Michael Chen, Joshua Clymer, Jai Dhyani, Elena Ericheva,
  Katharyn Garcia, Brian Goodrich, Nikola Jurkovic, Megan Kinniment, Aron
  Lajko, Seraphina Nix, Lucas Sato, William Saunders, Maksym Taran, Ben West,
  and Elizabeth Barnes.
\newblock {RE-Bench}: Evaluating frontier {AI} {R\&D} capabilities of language
  model agents against human experts.
\newblock In \emph{Proceedings of the 42nd International Conference on Machine
  Learning}, volume 267 of \emph{Proceedings of Machine Learning Research},
  pp.\  66772--66832. PMLR, 2025.

\bibitem[Yang et~al.(2024)Yang, Wang, Lu, Liu, Le, Zhou, and
  Chen]{yang2024opro}
Chengrun Yang, Xuezhi Wang, Yifeng Lu, Hanxiao Liu, Quoc~V. Le, Denny Zhou, and
  Xinyun Chen.
\newblock Large language models as optimizers.
\newblock In \emph{The Twelfth International Conference on Learning
  Representations (ICLR)}, 2024.

\end{thebibliography}
\bibliographystyle{tmlr}

\appendix

\section{Parameters of the search}
\label{app:params}

\begin{table}[H]
\centering\small
\caption{Search parameters (stage~1 had neither the shared pool nor the combiner, and otherwise used the same settings except where noted).}
\label{tab:params}
\begin{tabular}{@{}L{3.6cm}L{12.3cm}@{}}
\toprule
Component & Setting \\
\midrule
Length scale & $\ell=\sqrt{4\pi/N}$ \\
Replicas, temperatures & 3; $0.003$, $0.0173$, $0.1$ (geometric) \\
Moves (stages 2 and 3) & displacement of $k\sim\mathcal{U}\{6,\cdots,29\}$ neighbours, $\sigma\sim\mathcal{U}(0.1,0.35)\,\ell$, probability 0.55; rotation of $k\sim\mathcal{U}\{10,\cdots,39\}$ neighbours by $\theta\sim\mathcal{U}(-0.35,0.35)$, probability 0.35; displacement of $k\sim\mathcal{U}\{40,\cdots,89\}$ neighbours, $\sigma=0.08\,\ell$, probability 0.10 \\
Moves (stage 1) & first two move types with probabilities 0.65 and 0.35 \\
Relaxation & L-BFGS (SciPy implementation, no bounds), memory 30, gradient tolerance $10^{-6}$, relative function tolerance $10^{-13}$, at most 30{,}000 iterations, on unnormalised coordinates \\
Polishing & step $\alpha_t$ of \cref{sec:method} (the first Barzilai--Borwein step), capped at $10^{-2}$ and halved when $\Delta X_t^{\top}\Delta g_t\le0$; stop at $|g|_\infty<10^{-8}$ (stage~1 polishing: the lowest of three L-BFGS minimisations to gradient tolerance $10^{-9}$, two of them after Gaussian perturbations of standard deviation $10^{-4}$; released $N=1382$ configuration: $\alpha_t$ alternated with the second Barzilai--Borwein step $\Delta X_t^{\top}\Delta g_t/\lVert\Delta g_t\rVert^2$) \\
Polishing trigger & plain-sum energy below the pool best by $5\times10^{-6}$ (stage 1: below the baseline by $\max(2\times10^{-5},3\times10^{-11}E)$) \\
Pool novelty & energy differs by more than $2\times10^{-7}$ from that of every pool member \\
Reseeding & every 30 min, coldest replica from the pool best; other replicas from the four best pool configurations at start \\
Combiner & every 5 min; the ten best pool configurations below the baseline; aligned fraction $>0.9$; moved threshold $0.15\,\ell$; enlargement radius $1.6\,\ell$ \\
Candidate threshold & $10^{-5}$ below the baseline (the CCD energy recomputed from the posted coordinates; after stage~3, the polished baseline) \\
ET (controlled comparison; default afterwards) & stop a relaxation after 20 iterations if no point has moved by $0.2\,\ell$ from the starting minimum \\
Local relaxation (stage~3) & active set: the $\min\{\max\{2.5k+50,\,80\},\,260\}$ points nearest the move centre, where $k$ is the number of points moved; L-BFGS, memory 20, at most 600 iterations, gradient tolerance $10^{-7}$; hybrid arm re-relaxes fully if the local energy change exceeds $10^{-4}$ \\
Random seeds & stage~1: seeds 0 and 1 per size; stage~2: $1000+r$ for runs $r=0,1,2$; controlled comparison: $1000$ and $1001$ in every arm; probe of smaller sizes: 0, 1, 10, 11 and 12 per size \\
Energy kernel & C, $O(N^2)$, single-threaded; compensated summation for reported energies \\
Hardware & Apple M4 laptop (4 performance and 6 efficiency cores); up to nine single-core processes \\
\bottomrule
\end{tabular}
\end{table}

\section{Complete accounting of the searches}
\label{app:sweep}

\begin{table}[H]
\centering\small
\caption{Every size searched, with budget and outcome. ``Processes $\times$ h'' counts single-core processes: runs and, where marked, the combiner. Relaxations: L-BFGS minimisations after a move (in all $3N$ coordinates except in the local-relaxation runs of stage~3). $\Delta E$ values are rounded to four decimals (\cref{tab:main} gives those of the new configurations to seven decimals). In the Best $\Delta E$ column, a dash means that no configuration below the baseline was found (for $N=960$ the baseline is the challenge record from which the search started, $0.0449$ below the CCD value) or, in stage~3, that no size improved; in the other columns, a dash means that the quantity is not reported. Lowest distinct minimum: the lowest minimum found other than the starting configuration. Preliminary: the $N=960$ attempt (\cref{app:n960}). Stage~2 ran in two segments, of $1.6$ and $8$ hours; stage~3 is reported per size in \cref{tab:sweep}. The probe of smaller sizes was run on other hardware (per size, two runs on a laptop and three on a server); its budget is given in relaxations (five runs per size).}
\label{tab:accounting}
\begin{tabular}{@{}llrrlL{5.3cm}@{}}
\toprule
Stage & $N$ & Processes $\times$ h & Relaxations & Best $\Delta E$ & Notes \\
\midrule
Preliminary & 960 & $9\times10$ & $1.88\times10^6$ & -- & five strategies; about $80$ core-hours used \\
\midrule
\multirow{5}{*}{1 (screening)} & 1382 & $2\times2$ & 25{,}839 & $-0.0110$ & both runs below the CCD value, the first after 1.5 min \\
 & 2000 & $2\times2$ & 13{,}164 & -- & lowest distinct minimum $+0.0030$ \\
 & 2502 & $2\times2$ & 8{,}796 & $-0.0149$ & both runs below the CCD value, the first after 1.2 min \\
 & 3552 & $2\times2$ & 4{,}785 & -- & lowest distinct minimum $+0.0006$ \\
 & 2502 & $1\times2$ & -- & -- & symmetry-constrained ($C_2,C_3,C_6,D_3$); best $+0.459$ \\
\midrule
\multirow{6}{*}{2 (pool)} & 1382 & $2\times9.6$ & 101{,}646 & $-0.0110$ & no change \\
 & 2502 & $3\times9.6$ & 52{,}367 & $-0.0614$ & $-0.0511$ in the first segment \\
 & 2192 & $1\times9.6$ & 21{,}128 & -- & \\
 & 3182 & $1\times9.6$ & 11{,}358 & -- & \\
 & 4352 & $1\times9.6$ & 6{,}499 & -- & \\
 & all & $1\times9.6$ & -- & $-0.0614$ & combiner; 149 combinations, two of them before the relabelling fix; one new best \\
\midrule
\multirow{2}{*}{3 (time slices)} & 11 sizes & $8\times10$ & 203{,}520 & -- & local relaxation for $N\ge2000$ (\cref{sec:speedup}) \\
 & all & $1\times10$ & -- & -- & combiner; 240 combinations; nothing lower \\
\midrule
\makecell[l]{Controlled\\comparison} & 2502 & $8\times2$ & -- & $-0.0196$ & four arms, pass criterion fixed in advance (\cref{tab:control}) \\
\midrule
\multirow{4}{*}{\makecell[l]{Probe of\\smaller sizes}} & 900 & -- & 116{,}226 & -- & lowest distinct minimum $+0.0001$ \\
 & 912 & -- & 117{,}852 & $-0.0036$ & one of five runs (\cref{app:probe}) \\
 & 920 & -- & 113{,}817 & -- & lowest distinct minimum $+0.0004$ \\
 & 932 & -- & 109{,}965 & -- & lowest distinct minimum $+0.0019$ \\
\bottomrule
\end{tabular}
\end{table}

\begin{table}[H]
\centering\small
\caption{Stage~3 (time slices of 1.5--2 h cycling over 11 sizes, eight runs, about ten hours). Local relaxation was active for $N\ge2000$. No size improved; for $N=1382$ and $2502$ the starting point was the pool from stage~2.}
\label{tab:sweep}
\begin{tabular}{@{}lrrrrrrrrrrr@{}}
\toprule
$N$ & 1382 & 2000 & 2192 & 2502 & 3182 & 3552 & 4352 & 5792 & 5872 & 9911 & 9913 \\
\midrule
Core-hours & 7.5 & 7.9 & 7.5 & 7.5 & 6.9 & 7.0 & 7.0 & 8.0 & 7.0 & 7.0 & 6.5 \\
Relaxations ($10^3$) & 42.1 & 31.9 & 26.8 & 24.5 & 17.8 & 16.2 & 13.2 & 11.0 & 9.2 & 5.5 & 5.2 \\
Relaxation & full & local & local & local & local & local & local & local & local & local & local \\
\bottomrule
\end{tabular}
\end{table}

\section{Polishing dividends of all CCD entries}
\label{app:polish}

\begin{figure}[H]
\centering
\includegraphics[width=\linewidth]{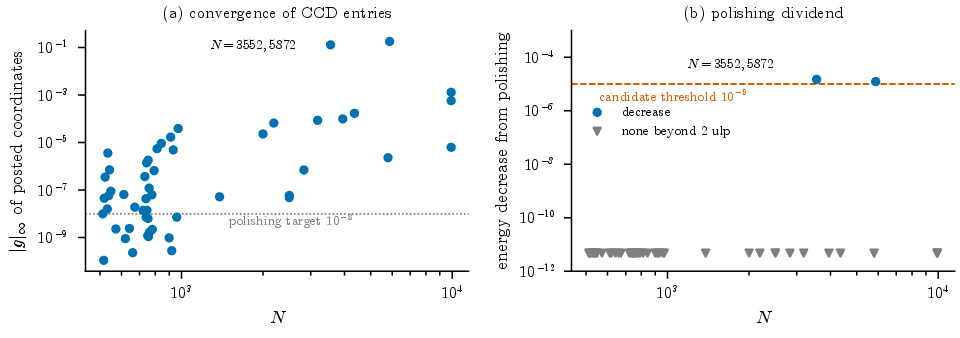}
\caption{Convergence of the 53 CCD entries. (a) $|g|_\infty$ of the posted coordinates. (b) Energy decrease obtained by polishing each CCD configuration (the iteration of \cref{sec:method}, to $|g|_\infty<10^{-8}$), relative to the energy recomputed from the posted coordinates; grey triangles, drawn at an arbitrary floor, mark entries whose energy changed by at most 2 ulp. Only $N=3552$ and $N=5872$ change by more than 2 ulp ($1.50\times10^{-5}$ and $1.23\times10^{-5}$). Dotted line in (a): the polishing target $|g|_\infty=10^{-8}$; dashed line in (b): the candidate threshold $10^{-5}$. CCD values agree with the recomputed energies to within $6\times10^{-8}$ (they are rounded to seven decimals); the values for every entry are provided as supplementary material.}
\label{fig:polish}
\end{figure}

\section{The preliminary attempt at \texorpdfstring{$N=960$}{N=960}}
\label{app:n960}

The project began as an entry to the Few Shots challenge \citep{fewshots2026}, whose problem p007, posted on 27 October 2026, is the Thomson problem for $N=960$, scored by an official verifier in exact integer arithmetic with the CCD configuration as baseline. The CCD lists a $T$-symmetric configuration with energy $444368.4373480$ for this size, set by \citet[Table~3]{amore2026efficient}, who lowered the previous value \citep{roshal2015} by $0.382$. Within a day of the problem's posting, two participants working with the Claude Code agent submitted lower configurations: one with $D_2$ symmetry ($444368.4040805$), found with a symmetry-preserving cut-and-splice genetic algorithm \citep{abreto2026p007}, and one with $C_3$ symmetry ($444368.3924062$), found by replica exchange in the $C_3$-symmetric subspace \citep{ronnie2025p007}, $0.0449$ below the CCD value. These are not our results and we make no claim about them; we report them because they are the closest prior use of the same tools on the CCD, and because they were our starting point.

Starting from the $C_3$ configuration, nine runs used five strategies for up to ten hours: replica-exchange basin hopping constrained to $C_3$ symmetry (the method behind the challenge record, as a control), $C_3$-preserving cut-and-splice recombination, unconstrained replica exchange, a targeted scan of scar compositions motivated by elasticity theory \citep{bausch2003}, and adding and removing a point. In about $80$ core-hours ($1.88\times10^6$ relaxations) none found a lower configuration: relative to the starting $C_3$ configuration, the symmetry-constrained runs stalled within minutes at $+0.0190$, and the lowest distinct minimum of the unconstrained runs was $+0.00328$, which short independent tests by the reviewing AI system also reached. We concluded that this record, freshly lowered by other participants, was a poor target for further search with our budget, and moved to the large-$N$ entries.

\section{The probe of smaller sizes}
\label{app:probe}

The probe ran the stage-1 protocol (stage-1 settings of \cref{tab:params}) with the project's search code, unchanged, on the four CCD entries without symmetry just below $N=960$; every run started from the CCD configuration. \Cref{tab:probe} lists the outcomes. At each size, the runs that did not go below the CCD value found the same lowest distinct minimum.

\begin{table}[H]
\centering\small
\caption{The probe of smaller sizes: five runs per size with the stage-1 protocol. Origin of the CCD value according to \citet[Tables~1 and~3]{amore2026efficient}: set by the October 2026 method, or set earlier and reproduced by this method. Relaxations: total over the five runs. Best $\Delta E$: $\Delta E$ of the lowest distinct minimum (\cref{tab:accounting}) found in the five runs.}
\label{tab:probe}
\begin{tabular}{@{}rlrrcr@{}}
\toprule
$N$ & Origin of CCD value & CCD value & Relaxations & \makecell[c]{Runs below\\CCD value} & Best $\Delta E$ \\
\midrule
900 & October 2026 method & $390084.7556201$ & 116{,}226 & 0 & $+0.0000591$ \\
912 & earlier, reproduced & $400657.6165027$ & 117{,}852 & 1 & $-0.0036186$ \\
920 & earlier, reproduced & $407784.8388324$ & 113{,}817 & 0 & $+0.0004066$ \\
932 & earlier, reproduced & $418594.2339500$ & 109{,}965 & 0 & $+0.0018764$ \\
\bottomrule
\end{tabular}
\end{table}

The checks of the $N=912$ configuration are reported with those of the other two in \cref{sec:results}.

\section{Reproducibility}
\label{app:repro}

\paragraph{Files.} The released coordinate files are identical to those analysed here; the leading digits of their hashes under the Secure Hash Algorithm with 256-bit digests are \texttt{19a0007473c235d2} ($N=2502$) and \texttt{8e2ee8176c6f1d46} ($N=1382$). The CCD coordinates were downloaded from \url{https://www-wales.ch.cam.ac.uk/~wales/CCD/Thomson/site/s_1/dat/}.

\paragraph{Certificates.} The rational coordinates used in the integer certificate \eqref{eq:cert} are provided for $N=912$, $1382$ and $2502$: each point is an integer quadruple $(a,b,c,d)$ with $a^2+b^2+c^2=d^2$, together with the energy bounds; anyone can re-check the bounds with exact integer arithmetic.

\paragraph{Scripts.} The supplementary scripts produce the integer certificates; the energies, gradients, Hessian spectra, defects and displacements; the polishing dividends of all CCD entries; the tests in \cref{sec:speedup}; the checks of the $N=912$ configuration; and \cref{fig:trajectory,fig:structure,fig:control,fig:polish}. They need Python with NumPy, SciPy and Matplotlib \citep{harris2020numpy,virtanen2020scipy,hunter2007matplotlib} and a C compiler.

\end{document}